\documentclass{article}

\usepackage{arxiv, times, natbib}

\usepackage{amsmath,amsfonts,bm}

\def\eqref#1{equation~\ref{#1}}

\def\1{\bm{1}}

\DeclareMathAlphabet{\mathsfit}{\encodingdefault}{\sfdefault}{m}{sl}
\SetMathAlphabet{\mathsfit}{bold}{\encodingdefault}{\sfdefault}{bx}{n}

\def\gC{{\mathcal{C}}}

\def\gJ{{\mathcal{J}}}

\def\gM{{\mathcal{M}}}
\def\gN{{\mathcal{N}}}

\newcommand{\E}{\mathbb{E}}

\newcommand{\R}{\mathbb{R}}

\usepackage{float}
\PassOptionsToPackage{hyphens}{url}
\usepackage{hyperref}
\usepackage{url}
\usepackage{graphicx}
\usepackage{algpseudocode}
\usepackage{algorithm}
\usepackage{tablefootnote}
\usepackage{color}
\usepackage{xcolor}
\usepackage{tikz}
\usetikzlibrary{arrows.meta, calc}
\usepackage{amsmath}
\usepackage{amsfonts}
\usepackage{amssymb}
\usepackage{tabularx}
\usepackage{multirow}
\usepackage{tcolorbox}
\usepackage{amsthm}
\usepackage{booktabs}
\usepackage{longtable}
\usepackage{adjustbox}
\graphicspath{{figures/}}

\usepackage{xcolor}
\usepackage{listings}

\lstdefinestyle{promptstyle}{
  basicstyle=\ttfamily\fontsize{7pt}{8pt}\selectfont,   
  frame=single,                 
  columns=fullflexible,
  keepspaces=true,
  showstringspaces=false,
  breaklines=true,              
  breakatwhitespace=false,      
  postbreak=\mbox{\textcolor{gray}{$\hookrightarrow$}\space}, 
  morecomment=[l]{\#},
  commentstyle=\bfseries\ttfamily,
  moredelim=[is][\color{black}]{\{}{\}},
}

\lstnewenvironment{prompt}[1][]{
  \lstset{style=promptstyle,#1}
}{}

\theoremstyle{definition}

\title{Intelligence from Learnable Novelty}

\author{Yanbo Zhang~$^{1}$ \quad Michael Levin~$^{1,2}$\thanks{Author of correspondence: \texttt{Michael.Levin@tufts.edu}}\\
~$^1$ Allen Discovery Center at Tufts University, Medford, MA, 02155, USA\\
~$^2$ Wyss Institute for Biologically Inspired Engineering at Harvard University,\\
~~~Boston, MA, 02115, USA
}

\begin{document}

\maketitle

\setcounter{footnote}{0}
\renewcommand{\thefootnote}{\fnsymbol{footnote}}
\begin{abstract}
    Intelligence appears under different names in different fields: as data compression in statistics and machine learning, as universal computation in dynamical systems, and as adaptive behavior in agents. Each field carries its own objective, and the two most influential drives often fail in mirror image: novelty search, which seeks surprise, is transfixed by a noisy television screen, while the free-energy principle, which avoids surprise, is most content in a dark room. Both failures have a single cause: each objective treats as one quantity the surprise a learner can convert into knowledge and the surprise it never can. Here we show that the learnable part of that information, which we call learnable novelty, yields the seemingly disparate projections of intelligence, and we give a closed-form estimator of it built on a cheap and differentiable reservoir computer. Used as a measure, with no supervision of any kind, the estimator recovers decades of complexity classification, ranking the Turing-complete rule~110 highest among the elementary cellular automata. Used as an objective, its gradient carries a neural cellular automaton from simple dynamics into a regime of solitons, the traveling, colliding structures by which rule~110 computes, as well as organizes the representation of an image encoder around the ten digit classes of MNIST, fully unsupervised: no label ever enters training. Handed to a reinforcement-learning agent as an intrinsic reward, it supplies the exploration that task rewards lack, improving on the task baseline in nine of ten environments and collapsing in none. Complexity generation, abstraction, and exploration, ordinarily pursued with unrelated objectives in separate fields, thus emerge from ascent on one differentiable quantity, and the projections of intelligence gain a common quantitative footing.
    \footnote{Code and reproduction materials: \url{https://github.com/Zhangyanbo/learnable-novelty}.}
\end{abstract}


\renewcommand{\thefootnote}{\arabic{footnote}}
\setcounter{footnote}{0}

\section{Introduction}

Few concepts are invoked across as many disciplines as intelligence, and few are theorized in as many incompatible ways. To statistics and machine learning it is extreme compression of data; to the study of complex systems it is the emergence of universal computation; in the interaction of an agent with its environment it is open-ended adaptive behavior. Each appearance has its own literature and its own objective function, and the literatures rarely meet. Here we show that these appearances follow from a single principle: the pursuit of learnable novelty.

Many creative processes unfold without a preset destination. Biological evolution has no fixed target, and scientific discovery often proceeds without knowing where it will lead. Both illustrate search in which the next direction cannot be specified in advance. \citet{lehman2011novelty} made this intuition operational as novelty search, which abandons objectives, rewards only behavior not seen before, and thereby escapes the deceptive local optima that trap goal-directed search. A parallel tradition holds that compression is intelligence: a model that compresses a data stream better has captured more of the mechanism that generated it, and generalizes better for it~\citep{hutter2005universal, deletang2024language}. In the same spirit, the free-energy principle~\citep{friston2010freeenergy} holds that a competent agent minimizes its cumulative surprise, keeping its world compressed. Once these ideas are turned into optimization objectives, however, they often fail, each in its own way. A learner that maximizes novelty is captured by a noisy television screen: unpredictable forever, hence forever novel, it teaches nothing~\citep{pathak2017curiosity, burda2019rnd}. A learner that minimizes surprise retreats into a dark room where nothing happens at all, because nothing is easier to predict~\citep{sun2020darkroom}. The two pathologies are mirror images with a common cause: both objectives conflate the total novelty in the data with the structure a bounded mind can absorb.

\begin{figure}[t!]
    \centering
    \input{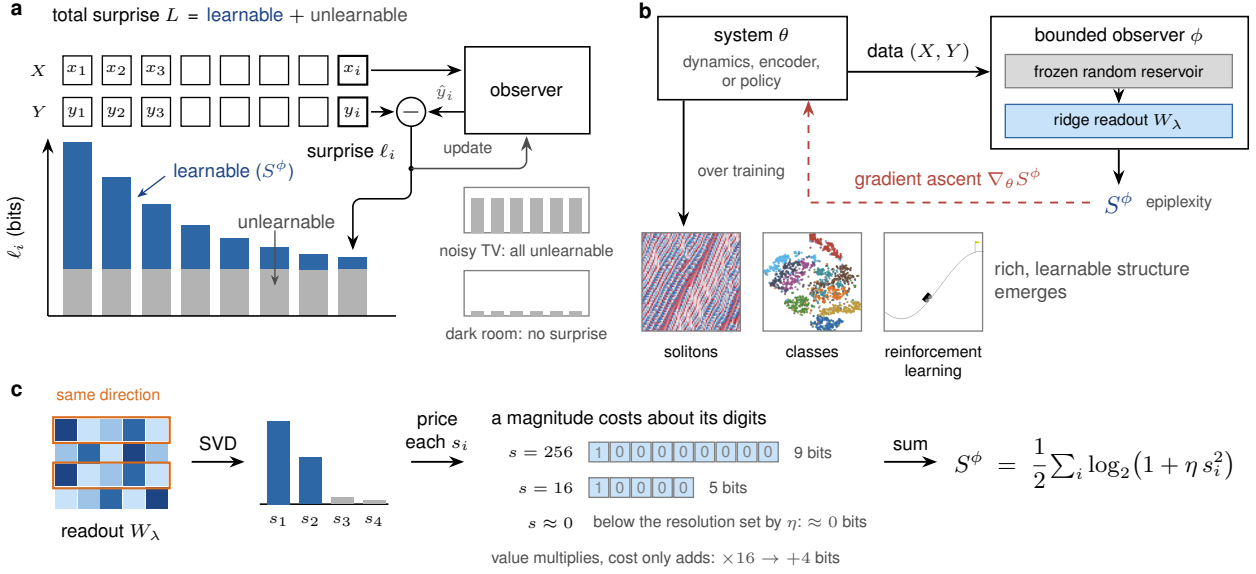}
    \caption{\textbf{(a)} An observer receives a stream one item at a time. It predicts each item before seeing it, is surprised by the difference, and updates itself, so as the stream's regularities are absorbed the surprise falls, though never below the floor of irreducible noise. The learnable part of the accumulated surprise, in blue, is the epiplexity $S^\phi$. A noisy TV is all noise, and a dark room offers no surprise at all. \textbf{(b)} The same quantity serves as an objective. The observed system produces data, the bounded observer condenses them into the single number $S^\phi$, and the gradient of that number flows back and reshapes the system. Driven by this alone, a cellular automaton develops solitons, an image encoder separates the digit classes, and an agent learns to explore. \textbf{(c)} Behind $S^\phi$ is the description length of what the observer has learned. Redundant directions merge and add nothing, and each remaining direction takes about as many bits as its magnitude has digits, so $S^\phi$ measures how much independent structure was learned rather than how large it is.}
    \label{fig:overview}
\end{figure}

Consider instead a learner of finite compute observing data one item at a time. Each item carries a quantum of surprise and an occasion to update the learner's model; as the stream accumulates, its regularities are internalized into the model while its irreducible randomness remains forever uncompressed. The surprise summed over the whole course is exactly the stream's minimum description length under that compute bound~\citep{dawid1984prequential, blier2018description, rissanen1978mdl, grunwald2007mdl, finzi2026epiplexity}. The total splits in two: the program length of the best model that finite compute can fit, and the residual that this model can never reduce (Figure~\ref{fig:overview}a). The first part, the structure a bounded mind genuinely carries away from the data, is the \emph{epiplexity} recently named by \citet{finzi2026epiplexity}, and it is what we mean by learnable novelty; Section~\ref{sec:novelty} states the decomposition formally. That work computes epiplexity as a measure of data already in hand; we read the same quantity as learnable novelty and give it a cheap, differentiable estimator, which turns it from a property to be measured into an objective a system can be driven to maximize. Seen through this decomposition, a noisy television is all residual, and a dark room contributes to neither part. Maximizing the learnable part alone therefore removes both pathologies at once.

What kind of system results when learnable novelty is maximized? For a bounded learner to keep extracting rich structure, the system it learns from must sit at the boundary of order and chaos: too ordered and there is nothing to learn, too chaotic and nothing is learnable. A dynamical system optimized for learnable novelty should therefore approach general-purpose computation, since only a system that can compute arbitrarily can keep producing learnable structure without bound~\citep{wolfram1984universality, langton1990computation, cook2004universality}. A representation optimized for learnable novelty must shed redundancy and emit as many mutually distinguishable, recoverable responses as the data support, so category-like structure should emerge without supervision. And an agent whose policy maximizes the learnable novelty of its own future is an agent whose death or stagnation terminates that extraction: it is driven to avoid termination, to preserve its capacity to act, and to keep its exchange with the environment rich. This sharpens the causal-entropic~\citep{wissner2013causal} and empowerment~\citep{klyubin2005empowerment, salge2014empowerment} accounts of intelligent behavior: the reachable futures must be not merely diverse but learnable.

Testing these predictions means evaluating learnable novelty inside an optimization loop, and there its cost becomes prohibitive: the definition of epiplexity calls for a search over bounded-compute models, and the original work trains a neural network for every system scored~\citep{finzi2026epiplexity}. But the definition does not fix what the bounded learner is, only that it learns. A program class qualifies, a neural network qualifies, and so does a reservoir computer~\citep{jaeger2004harnessing, maass2002real}, in which all learnable capacity resides in a linear readout whose optimum is the closed-form solution of a ridge regression. With the reservoir as the learner, the score becomes cheap, deterministic, and differentiable in whatever produced the data: learnable novelty becomes not merely measurable but directly optimizable (Figure~\ref{fig:overview}b).

Used as a measure, with no training of any kind, it reproduces complexity rankings accumulated over decades of study: across the elementary cellular automata it places the Turing-complete rule~110~\citep{cook2004universality} at the top. Used as an objective, its gradient drives structure into existence. With no supervisory signal, it carries a one-dimensional neural cellular automaton~\citep{mordvintsev2020growing} from simple behavior into a system of complex solitons, coherent traveling structures that collide and interact. The same gradient carries a randomly initialized encoder of MNIST~\citep{lecun1998mnist} into a representation in which the digit classes separate, with no label ever entering training. Used as a reinforcement-learning reward, it drives an agent to explore novel, richer behavior on its own, reaching higher return faster in sparse-reward, deceptive environments.

\section{From Novelty to Epiplexity}
\label{sec:novelty}

Novelty search made the pursuit of the unseen operational but left \emph{novelty} itself undefined: it hand-picks a behavior descriptor and rewards distance in that chosen space~\citep{lehman2011novelty}. A measure of novelty should instead be intrinsic to the data, and the natural raw material is surprise. Consider transmitting a sequence~$Y = (y_1, \dots, y_N)$ to a receiver who already holds the corresponding inputs~$X = (x_1, \dots, x_N)$. The data arrive one symbol at a time, and before each arrival the receiver predicts the symbol from what it has already seen, $p(y_i \mid y_{<i}, X)$. When~$y_i$ is revealed, the receiver pays a surprise cost~$\ell_i = -\log_2 p(y_i \mid y_{<i}, X)$ and updates its predictor on the new datum. Summed over the sequence, the chain rule collapses the total cost to a single quantity,
\begin{equation}
    L \;=\; \sum_{i=1}^{N} \ell_i \;=\; -\log_2 \prod_{i=1}^{N} p(y_i \mid y_{<i}, X) \;=\; -\log_2 p(Y \mid X),
    \label{eq:prequential}
\end{equation}
the prequential description length~\citep{dawid1984prequential, blier2018description}: the number of bits the receiver pays before~$X$ determines~$Y$. Cumulative surprise is a natural gauge of novelty: the more an observer who holds~$X$ is surprised in the course of observing~$Y$, the more novel the relationship between them. Rewarding~$L$ in place of distance in a hand-chosen behavior space preserves the drive toward the unseen while removing the arbitrary descriptor.

Read this way, two established programs occupy the two extremes of one quantity. Novelty search maximizes~$L$: high cumulative surprise marks behavior not seen before, and chasing it outperforms goal-directed search on deceptive problems~\citep{lehman2011novelty}. The free-energy principle minimizes~$L$: low cumulative surprise marks an internal model that fits its inputs, and a surprise-minimizing agent is read as self-organizing~\citep{friston2010freeenergy}. Each fails, but for opposite reasons. A noisy television maximizes~$L$ per unit time because every frame is independent random noise, so a surprise-seeking agent parks in front of it and learns nothing~\citep{pathak2017curiosity, burda2019rnd}. A dark room minimizes~$L$ because every input repeats the last, so a surprise-minimizing agent prefers it to any structured environment~\citep{sun2020darkroom}. The two failures share a root: $L$ is a sum of two distinct components, the surprise that can be learned and the surprise that cannot. An objective that sees only the sum can be driven to either extreme by moving one component alone.

The minimum description length principle makes the two components precise~\citep{rissanen1978mdl, grunwald2007mdl, solomonoff1964formal}. Among all models that could carry~$Y$ to a receiver holding~$X$, the minimizing one splits the transmission into a description of the model and the residual the model cannot account for,
\begin{equation}
    L \;\approx\; \min_{M \in \gM} \Big[\,|M| - \log_2 p(Y \mid X, M)\,\Big],
    \label{eq:mdl}
\end{equation}
where the approximation errs only in lower-order terms~\citep{grunwald2007mdl}. The first term, the program length~$|M|$, is the part of the relationship an observer can \emph{learn} and reuse on future inputs. The second is the part that, at the minimum, is unlearnable: not because no model could compress it further, but because any model that tried would only lengthen the total transmission. This decomposition identifies the flaw in each of the two earlier programs. Novelty search is dominated in the extreme by the residual term: the noisy television contributes nothing to~$|M|$. The free-energy principle, in minimizing prediction error, compresses~$|M|$ along with it, and is therefore drawn to the dark room. Both flaws point to the same conclusion: the quantity worth pursuing is~$|M|$ alone.

Equation~\eqref{eq:mdl} is uncomputable in general, since no search can visit every program in~$\gM$, and real observers hold only finite compute; we therefore consider only a finite set of halting programs~$\gM_\phi$, given by a fixed bounded observer~$\phi$. The computable restriction of the MDL problem is
\begin{equation}
    \begin{aligned}
        L^\phi(Y \mid X) &= \min_{M \in \gM_\phi} \Big[\,|M| - \log_2 p(Y \mid X, M)\,\Big], \\
        M_\phi^*(Y \mid X) &= \arg\min_{M \in \gM_\phi} \Big[\,|M| - \log_2 p(Y \mid X, M)\,\Big].
    \end{aligned}
\end{equation}
Here we define~$|M_\phi^*|$ as the learnable novelty; it is precisely the \emph{epiplexity} of~\citet{finzi2026epiplexity}, the structure a finite-compute learner can actually extract from the data:
\begin{equation}
    S^\phi(Y \mid X) \;=\; |M_\phi^*(Y \mid X)|.
    \label{eq:epiplexity-def}
\end{equation}
By contrast, novelty search must postulate its behavior space by hand, whereas learnable novelty does not: it needs only a bounded observer to decide what counts as novel, and that boundedness is why both the noisy television and the dark room contribute zero to~$|M_\phi^*|$.

Epiplexity measures data that are given; learnable novelty is that quantity pursued as a goal, and it asks that the process generating the data be optimized. When the optimized object is the \emph{dynamics} of the data, maximizing~$S^\phi$ pushes it into the regime that emits the most learnable structure: trivial dynamics offer nothing to learn and fully chaotic dynamics nothing learnable, so the maximum lies between order and chaos~\citep{langton1990computation, packard1988adaptation}, where some systems can sustain universal computation~\citep{wolfram1984universality, cook2004universality}. When the optimized object is a \emph{representation}~$x \mapsto z$ of static data, maximization forces a bounded observer to recover from~$x$ as many independent, learnable distinctions in~$z$ as possible while shedding redundant information, and this is precisely compression: the code then organizes into clusters, the structure long associated with compression as understanding~\citep{hutter2005universal, deletang2024language}. When the optimized object is the \emph{policy} of an embodied agent, maximizing the learnable novelty of its reachable futures selects the actions that lead to rich future states; conversely, actions that terminate the agent or collapse its options remove reachable futures and lower the score, while actions that leave the environment in a richer state preserve them. This is the causal-entropic account of intelligent behavior~\citep{wissner2013causal}; learnable novelty goes one step further and asks that the reachable futures be not merely diverse but learnable. Section~\ref{sec:experiments} tests all three predictions.

\section{A Closed-Form Estimator of Epiplexity}
\label{sec:method}

Evaluating epiplexity requires finding the optimum in the bounded model space~$\gM_\phi$, which amounts to a full training run for every system scored; the original epiplexity work does exactly this~\citep{finzi2026epiplexity}. This is not only computationally expensive but also opaque to gradients: nothing backpropagates through such a training run to the system being scored, so epiplexity evaluated this way serves as a measure but not as an objective.

Yet the definition does not prescribe a particular learner architecture; it only requires a system with bounded learning capacity. A reservoir computer~\citep{jaeger2004harnessing, maass2002real} fits this requirement exactly: a fixed, randomly initialized nonlinear feature map followed by a single linear readout layer. With the reservoir as the bounded learner, finding the optimal model reduces to solving a ridge regression in closed form, making epiplexity cheap to evaluate, deterministic, and fully differentiable.

Concretely, let~$\phi$ be an untrained, fixed random nonlinear map. Applied to~$N$ inputs~$X$, it yields a feature matrix~$H = \phi(X) \in \R^{N \times m}$, where~$m$ is the reservoir feature dimension, with targets~$Y \in \R^{N \times D}$ (the construction of~$\phi$ is given in Appendix~\ref{app:criticality}). All learnable capacity resides in the linear readout matrix~$W \in \R^{m \times D}$, so~$\gM_\phi$ can be defined as~$\phi$ followed by every possible linear readout. In other words, $W$ is a program defined on the reservoir~$\phi$, and computing the model's program length reduces to computing the description length of the linear operator~$W$.

Under the minimum description length (MDL) framework~\citep{rissanen1978mdl, grunwald2007mdl}, the optimal readout minimizes the total description length: a residual part plus a weight part~$\gC(W, \phi)$. Assuming Gaussian residual noise~$Y = H W + \epsilon$, the residual part in bits equals a scaled mean squared error, and the total description length is
\begin{equation}
    L(W) \;=\; \frac{\|Y - H W\|_F^2}{2\sigma^2 \ln 2} + \gC(W, \phi).
    \label{eq:total-code}
\end{equation}

For the weight part we take a spectral description length based on the singular values~$s_i(W)$ of the readout:
\begin{equation}
    \gC_{\mathrm{spec}}(W)
    \;=\;
    \alpha \log_2 \det\!\big(I_m + \eta\, W W^\top\big)
    \;=\;
    \alpha \sum_i \log_2\!\big(1 + \eta\, s_i(W)^2\big),
    \label{eq:spectral-code}
\end{equation}
where~$\eta$ is a resolution parameter and~$\alpha$ is an overall scale on the description length. Since~$\alpha$ changes neither the ranking between systems nor the direction of the gradient, all experiments fix~$\alpha = 1/2$ and set~$\eta = 1$, except the MNIST encoder, which uses~$\eta = 30$ (Table~\ref{tab:reservoirs}). This log-determinant form follows the MDL principle that model complexity is measured by description length, matches the coding-rate objectives used in representation learning~\citep{rissanen1978mdl, grunwald2007mdl, yu2020mcr2}, and can also be derived from a hierarchical Gaussian prior (Appendix~\ref{app:matrix-t}). Its information-theoretic advantage is that scaling adds only logarithmic cost, and when readout directions coincide or are highly redundant they contribute no new independent singular value and incur only a logarithmically small additional cost, in compliance with the compression principle of algorithmic information theory (Figure~\ref{fig:overview}c).

With~\eqref{eq:spectral-code} as the weight part, however, the total description length~\eqref{eq:total-code} has no closed-form minimizer: solving it requires an inner iterative optimization, which raises the cost of every evaluation and obstructs the gradients we want to pass through~$W$ to the system being scored. We therefore approximate the minimization by a ridge regression~\citep{tikhonov1977regularization}; pricing weights by an $L_2$ penalty on MDL grounds is a classical move in machine learning~\citep{hinton1993keeping}, and a recent result relates the minimum weight norm of a fixed-precision network to the Kolmogorov complexity of the string it generates~\citep{musat2026weightnorm}. A Taylor expansion shows the two are close: for small~$W$, $\log_2 \det\!\big(I_m + \eta\, W W^\top\big) \approx \eta\, \|W\|_F^2 / \ln 2$, so the spectral description length itself reduces to the quadratic penalty of the ridge, and the residual scale~$\sigma^2$ merges with~$\alpha$ and~$\eta$ into a single ridge parameter~$\lambda$; the ridge's own shrinkage in turn keeps the readout in the small-norm regime where the expansion is accurate. To keep the features on a consistent scale, we standardize each feature column before solving the ridge problem; the target is centered and divided by a fixed scale factor~$u_Y$ posited in advance:
\begin{equation}
    \tilde{H}_{\cdot c} = \frac{H_{\cdot c} - \mu_c}{\hat{\sigma}_c\,\sqrt{m}},
    \qquad
    \tilde{Y} = \frac{Y - \mu_Y}{u_Y},
    \label{eq:normalization}
\end{equation}
where $\mu_c$ and~$\hat{\sigma}_c$ are the empirical mean and standard deviation of feature column~$c$, while~$\mu_Y$ and~$u_Y$ are the target centering and scale constants used by the estimator instance. The target is divided by~$u_Y$ rather than by its own empirical standard deviation because the target's magnitude itself carries information: a larger-magnitude target demands a larger readout, which the spectral description length~\eqref{eq:spectral-code} prices at more bits, logarithmically in its scale; $u_Y$ thus acts as the measurement precision posited for the target, the unit relative to which readout magnitude and residual are priced. The~$\sqrt{m}$ factor keeps the output scale of a random readout with i.i.d.\ standard normal entries invariant across feature widths.

The optimal readout can then be written in closed form:
\begin{equation}
    W_\lambda
    \;=\;
    \arg\min_W \|\tilde{Y} - \tilde{H} W\|_F^2 + \lambda \|W\|_F^2
    \;=\;
    (\tilde{H}^\top \tilde{H} + \lambda I_m)^{-1} \tilde{H}^\top \tilde{Y}.
    \label{eq:ridge-surrogate}
\end{equation}

Substituting~$W_\lambda$ into the spectral description length gives the complete reservoir-based closed-form estimator of epiplexity:
\begin{equation}
    \begin{aligned}
        S^\phi(Y \mid X) &= \frac{1}{2} \log_2 \det\!\big(I_m + \eta\, W_\lambda W_\lambda^\top\big)\\
        &= \frac{1}{2} \sum_i \log_2\!\big(1 + \eta\, s_i(W_\lambda)^2\big), \\
        \text{where }W_\lambda &\;=\; (\tilde{H}^\top \tilde{H} + \lambda I_m)^{-1} \tilde{H}^\top \tilde{Y}.
    \end{aligned}
    \label{eq:rc-score}
\end{equation}
Through this construction, an expensive bounded-model search is compressed into a closed-form ridge solve. The optimum is unique, and~$S^\phi$ is differentiable in~$(X, Y)$. In practice, for reasons of numerical stability, we never solve the normal equations of~\eqref{eq:ridge-surrogate} directly; an algebraically equivalent, better-conditioned least-squares solve computes the same~$W_\lambda$ and remains differentiable (Appendix~\ref{app:ridge-solve}). Appendix~\ref{app:ridge-surrogate} also minimizes the total description length~\eqref{eq:total-code} directly, without the approximation: the exact solution extracts more bits from the same data but ranks systems almost identically to the ridge readout.

\section{Experiments}
\label{sec:experiments}

Intelligent behavior often appears in different guises in different systems, and we propose that these are all the product of maximizing learnable novelty. We therefore test, in turn, nonlinear dynamical systems, representation learning, and reinforcement-learning tasks. Architectures, sampling procedures, and optimization settings for all experiments are included in Appendix~\ref{app:architectures}.

\subsection{Dynamical systems}
\label{sec:experiments-dynamics}

\begin{figure}[t!]
    \centering
    \includegraphics[width=\textwidth]{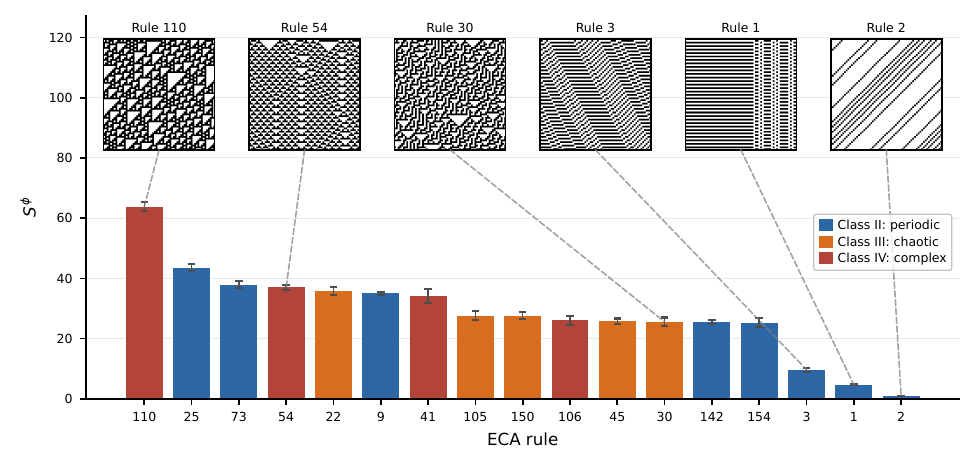}
    \caption{$S^\phi$ over all 88 locally unique elementary cellular automata (top fourteen shown, together with the reference rules). Bars are colored by Wolfram class~\citep{wolfram2002new}: II periodic, III chaotic, IV complex (legend); insets show the space-time diagrams (time downward) of the six reference rules 1, 2, 3, 30, 54, and 110. Each bar is the mean over ten independent draws of the random reservoir and the input ensemble, and error bars show one standard deviation. Rule 110 (Turing-complete, class IV) is the clear maximum over the whole space; at the other end, the rules whose sampled attractor dynamics are constant score exactly zero (none appears in the figure), and the near-trivial periodic rules 1 and 2 are the lowest-scoring rules shown.}
    \label{fig:eca-combined}
\end{figure}

Elementary cellular automata (ECA) are among the simplest nonlinear dynamical systems: a one-dimensional binary lattice in which each site determines its next state from its own state and those of its nearest neighbors. Each ECA rule takes only 8 bits to describe, yet within these minimal rules lies the potential for emergent complex computation.
Different ECA rules produce a rich range of behaviors (simple, chaotic, and complex), and among Wolfram's elementary rules, rule~110 is the only one currently proven Turing-complete~\citep{cook2004universality}. ECA are therefore the ideal testbed for asking what kind of dynamics maximizes learnable novelty.

For the discrete, finite-space elementary cellular automata we perform an exhaustive evaluation. After removing the various symmetries, we select 88 independent rules; for each we sample stationary states as the initial condition~$X$ on a width-$64$ ring and stack the next~$32$ one-step evolutions into the target~$Y$, scoring the map with~\eqref{eq:rc-score} through a convolutional reservoir. Rule~110 ranks highest in the entire rule space, the lowest scores, exactly zero, go to the rules whose sampled attractor dynamics are constant, the near-trivial periodic rules 1 and 2 score low, and the chaotic rule 30 lies in between, below the complex rule~54 (Figure~\ref{fig:eca-combined}). As a pure measurement tool, the closed-form estimator reproduces the classical complexity classification without any supervision, and rule~110's top rank is robust to the estimator's hyperparameters (Appendix~\ref{app:eca-robustness}). The same method, with the reservoir matched to the data geometry, also reproduces the complexity ordering of continuous-time dynamical systems (Appendix~\ref{app:observer-extra}), confirming the estimator's validity as a measure.

\begin{figure}[t!]
    \centering
    \includegraphics[width=\textwidth]{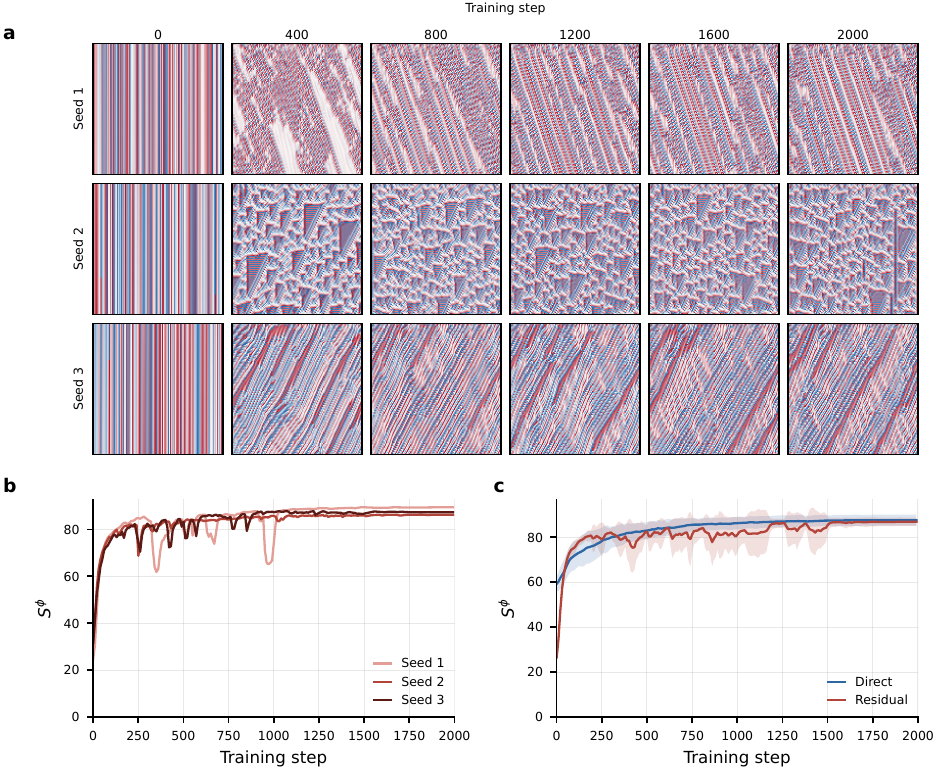}
    \caption{Inverse design of a one-dimensional NCA by gradient ascent on a single epiplexity scalar. \textbf{(a)} Space--time diagrams of the learned rule (time downward) for three random seeds: one row per seed, one column per training step from $0$ to $2000$. Every seed develops complex solitons from a simple initial rule: localized structures that travel at fixed velocity and interact on collision. \textbf{(b)} $S^\phi$ over training for each of the three seeds individually. \textbf{(c)} $S^\phi$ for both update rules, each drawn as the mean (line) $\pm$ one standard deviation (shaded band) over nine independent seeds.}
    \label{fig:nca-snapshots}
\end{figure}

To further test what happens when this learnable novelty is directly optimized, we turn to the continuous, two-channel neural cellular automata (NCA)~\citep{mordvintsev2020growing}. The local update~$G_\theta$ applies a learnable convolutional map~$g_\theta$ and renormalizes each site; we test it in two forms, a direct update
\begin{equation}
    x_{t+1} = G_\theta(x_t) = \operatorname{normalize}\big[g_\theta(x_t)\big],
\end{equation}
and a residual update~$G_\theta(x_t) = \operatorname{normalize}\big[x_t + g_\theta(x_t)\big]$ that adds a skip connection around~$g_\theta$; here~$\operatorname{normalize}$ fixes the two-channel vector at each site to unit length. Because~$G_\theta$ is differentiable, learnable novelty can be maximized by gradient ascent on~$\theta$ directly. At each training step, a batch of initial states is sampled and run forward for~$32$ gradient-free burn-in steps to reach a steady state, yielding~$X$; the dynamics are then rolled out differentially and the target stacks the next~$\tau$ states,
\begin{equation}
    Y_\theta(X,\tau) = \big(G_\theta(X), G_\theta^{2}(X), \dots, G_\theta^{\tau}(X)\big).
\end{equation}
Figure~\ref{fig:nca-snapshots} uses~$\tau=8$, scored by the frozen reservoir estimator, and the objective is simply
\begin{equation}
    \max_\theta \;\; S^\phi(Y_\theta(X,\tau) \mid X).
\end{equation}

As the score rises, the initially simple rule spontaneously develops solitons: localized coherent structures that travel at fixed velocity and interact on collision (Figure~\ref{fig:nca-snapshots}a; Appendix~\ref{app:inverse-extra}). Solitons are often a hallmark of a system's capacity for information transmission and combination: rule~110 uses them to carry and combine information. The regime is insensitive to the exact form of the local update: the direct and residual normalized updates climb to the same epiplexity band over nine seeds (Figure~\ref{fig:nca-snapshots}c) and both develop solitons at every seed (Figure~\ref{fig:nca-variants}). That gradient ascent reliably finds this regime is no accident: universal computation is where learnable novelty is unbounded (a system that can compute arbitrarily keeps producing structure a learner has not yet absorbed), so climbing learnable novelty naturally pushes the dynamics toward the edge of order and chaos, where coherent structures such as solitons live. Solitons suit the bounded observer particularly well: a structure propagating at fixed velocity transforms predictably from step to step, a regularity the linear readout captures cheaply, while its collisions keep producing configurations the readout has not yet absorbed.

\subsection{Representation learning}
\label{sec:experiments-encoder}

\begin{figure}[t!]
    \centering
    \includegraphics[width=\textwidth]{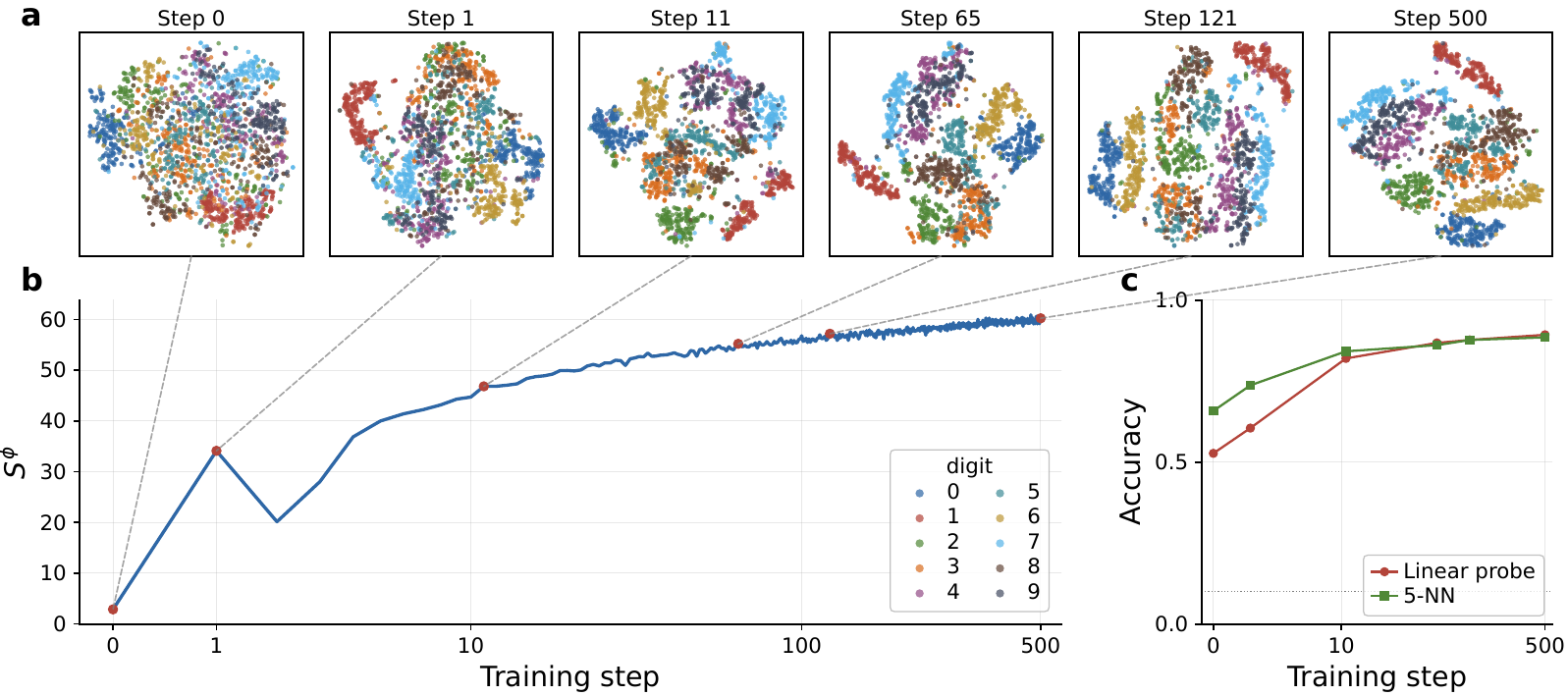}
    \caption{Unsupervised MNIST encoder trained solely to maximize reservoir epiplexity. \textbf{(a)} Two-dimensional t-SNE projections of the representation at six training checkpoints; colors indicate the held-out digit label, used for visualization only. \textbf{(b)} The $S^\phi$ training curve, with dashed connectors marking the checkpoint each projection is taken from. \textbf{(c)} Accuracy with which a linear probe and a $5$-nearest-neighbor classifier recover the digit from the code at the same six checkpoints (chance~$0.1$, dotted); both rise with the epiplexity, the linear probe from~$0.53$ to~$0.89$ and the $5$-nearest-neighbor from~$0.66$ to~$0.89$. Compact per-digit regions emerge progressively as the epiplexity rises, even though no class label ever enters training.}
    \label{fig:mnist-projections}
\end{figure}

If the object being optimized is a representation map, maximizing its learnable novelty should force the encoder to shed redundancy and spontaneously organize around the latent categories of the data. We test this hypothesis on MNIST~\citep{lecun1998mnist}. A trainable encoder~$E_\theta$ maps an image~$x \in X$ to a~$64$-dimensional feature vector~$z = E_\theta(x) \in \R^D$, normalized to unit length, and is trained to maximize the learnable novelty of its own code against a fixed random MLP reservoir (Appendix~\ref{app:inverse-extra}); no class label enters at any stage.

Here we maximize~$S^\phi(Z = E_\theta(X) \mid X)$. After training with no class labels whatsoever, two-dimensional t-SNE projections~\citep{maaten2008tsne} of the code at successive checkpoints show the initially entangled distribution progressively converging into clusters largely separated by digit class (Figure~\ref{fig:mnist-projections}). A linear probe~\citep{alain2016probes} and a $5$-nearest-neighbor classifier confirm this quantitatively: the accuracy with which each recovers the digit from the code rises together with the epiplexity, reaching $0.89$ for both probes by the end of training (Figure~\ref{fig:mnist-projections}c). The resulting representation quality is stable to moderate one-at-a-time changes of the training and estimator settings (Appendix~\ref{app:inverse-extra}).

Under this objective, if the encoder merely represents highly redundant information in the data, such as the black background shared by all MNIST samples, this information barely varies across samples and contributes nothing new from the standpoint of learnable novelty. To maximize the score, the encoder must capture the features that create substantive differences between inputs. Maximizing learnable novelty is therefore implicit data compression: it forces the encoder to shed redundancy and retain only the most discriminative factors of the data. On MNIST, the dominant such factor is the digit itself. The categorical structure of the representation space emerges spontaneously in the pursuit of this single metric.

This spontaneous cluster structure can also be read through the information bottleneck~\citep{tishby1999bottleneck}. For an input~$X$, a relevance variable~$Y$, and an intermediate representation~$Z$, the bottleneck minimizes~$I(X;Z) - \beta\, I(Y;Z)$ over encoding distributions: it keeps the information in~$Z$ that bears on~$Y$ and compresses the rest. The bottleneck needs the external variable~$Y$ to decide which bits are worth keeping; in our objective that criterion is supplied implicitly by the decodability of the bounded observer: whatever structure the reservoir's short readout can recover counts as relevant. Ridge shrinkage biases this criterion toward functions of low norm in the random-feature kernel space, functions that vary smoothly along the data manifold, and among these the class is the factor carrying the most discriminative information (the cluster assumption of semi-supervised learning~\citep{chapelle2006semi}); this experiment sets~$\lambda = 3$, far above the other experiments (Table~\ref{tab:reservoirs}), precisely so that the relevance criterion admits only this smooth structure. The role of the compression term~$I(X;Z)$ falls to the structural bottleneck: the code dimension is finite, and the spectral description length is nearly insensitive to repeated readout directions, so redundancy is squeezed out of the code.

\subsection{Reinforcement learning}
\label{sec:experiments-rl}

For an agent situated in an environment, pursuing learnable novelty should steer it away from dead ends and sustain the richness of its behavior. The drive is a relative of empowerment~\citep{klyubin2005empowerment, salge2014empowerment}, which maximizes the channel capacity from actions to future states and so rewards control over the future; learnable novelty asks in addition that those futures be learnable by the bounded observer. Because gradients cannot propagate through the environment, we hand the same closed-form score to a standard PPO algorithm~\citep{schulman2017ppo, raffin2021stable} as an intrinsic reward, and compare four conditions on each task: the environment's task reward alone, the trajectory's epiplexity alone (the agent never sees the task), the task reward plus a small epiplexity bonus, and a state-magnitude control introduced below.

From the observation trajectory~$o_0, o_1, \dots$, at each step the map from~$x = o_t$ to the future window~$y = (o_{t+1}, \dots, o_{t+\tau})$ is scored jointly by a frozen reservoir estimator of the same construction. To keep the reward on coherent novelty rather than noise or simple behavior, we adopt two fixed settings: the reservoir is kept small, so it fits a structured trajectory but not a chaotic one and the reward favors regular motion; and the window horizon~$\tau$ is matched to the characteristic time over which the state changes appreciably. To make the reward dense, we define the per-step bonus as the increment the latest state adds to the trajectory's epiplexity, so that in the mixed condition the agent maximizes $r_t = r_t^{\mathrm{task}} + \beta\,\big(S^\phi_t - S^\phi_{t-1}\big)$, with the weight~$\beta$ calibrated per environment (Appendix~\ref{app:rl}); where the task reward is flat, the increment supplies the gradient that drives the agent to explore new behavior.

\begin{table}[tb]
    \centering
    \caption{Epiplexity as a reinforcement-learning reward across ten tasks, mean $\pm$ standard deviation over ten seeds at $600{,}000$ steps. Columns: the return under the task reward alone; the trajectory's epiplexity alone (the agent never sees the task); the task reward plus a state-magnitude bonus ($\|o_t\|^2$, the squared norm of the input-normalized state); and the task reward plus the epiplexity bonus. \textbf{Bold} marks a run that improves on the task-reward baseline: the epiplexity bonus does so on every task but Walker2d and collapses on none (on Walker2d, epiplexity alone in fact exceeds the task reward), whereas the magnitude control falls below the baseline on Hopper, Walker2d, and LunarLander.}
    \label{tab:rl}
    \small
    \setlength{\tabcolsep}{4.5pt}
    \begin{tabular}{lrrrr}
        \toprule
        Task & Task reward & Epiplexity only & Task $+$ magnitude & Task $+$ epiplexity \\
        \midrule
        \multicolumn{5}{l}{\emph{Sparse / deceptive classic control}} \\
        \quad Acrobot              & $-167 \pm 166$ & $-500 \pm 0$  & $\mathbf{-89 \pm 4}$   & $\mathbf{-83 \pm 2}$    \\
        \quad MountainCarContinuous & $28 \pm 43$    & $-97 \pm 4$   & $\mathbf{93 \pm 1}$    & $\mathbf{93 \pm 1}$     \\
        \multicolumn{5}{l}{\emph{MuJoCo / Box2D locomotion}} \\
        \quad Hopper               & $1879 \pm 325$ & $1006 \pm 21$ & $516 \pm 128$          & $\mathbf{2192 \pm 270}$ \\
        \quad BipedalWalker        & $125 \pm 74$   & $-59 \pm 42$  & $\mathbf{146 \pm 51}$  & $\mathbf{151 \pm 49}$   \\
        \quad HalfCheetah          & $362 \pm 201$  & $-181 \pm 244$ & $\mathbf{623 \pm 212}$ & $\mathbf{393 \pm 260}$  \\
        \quad Walker2d             & $296 \pm 45$   & $\mathbf{327 \pm 45}$  & $294 \pm 32$           & $285 \pm 41$            \\
        \quad Swimmer              & $181 \pm 76$   & $-12 \pm 26$  & $\mathbf{194 \pm 76}$  & $\mathbf{206 \pm 86}$   \\
        \multicolumn{5}{l}{\emph{Sparse navigation}} \\
        \quad PointMaze            & $229 \pm 77$   & $6 \pm 3$     & $\mathbf{242 \pm 79}$  & $\mathbf{256 \pm 22}$   \\
        \multicolumn{5}{l}{\emph{Dense-reward control}} \\
        \quad LunarLander          & $169 \pm 74$   & $-438 \pm 393$ & $-171 \pm 35$          & $\mathbf{208 \pm 25}$   \\
        \quad Pendulum             & $-1044 \pm 42$ & $-1092 \pm 56$ & $\mathbf{-1016 \pm 21}$ & $\mathbf{-987 \pm 23}$  \\
        \bottomrule
    \end{tabular}
\end{table}

Across ten environments spanning classic control, MuJoCo locomotion, and maze navigation, the epiplexity bonus improves on the task reward on nine tasks (Table~\ref{tab:rl}). On the sparse and deceptive-reward tasks (Acrobot and MountainCarContinuous), the task reward alone yields returns that vary widely across seeds, with some seeds failing to converge; with the epiplexity bonus, all seeds solve the task and the across-seed standard deviation drops sharply. It supplies the exploration drive the task reward lacks (driving the cart back and forth to build momentum, pumping the acrobot up to its bar), turning unreliable exploration into reliable solving. On the higher-dimensional locomotion tasks the bonus also lifts the return, most on Hopper. On sparse maze navigation (PointMaze) the bonus likewise raises the return and sharply cuts its across-seed variance, from~$77$ to~$22$.

To verify that the agent is not merely cheating by chasing large state values (accumulating extreme coordinates in the physics engine), we introduce state magnitude as a control reward: the squared norm of the input-normalized state, calibrated identically (Table~\ref{tab:rl}, magnitude column). When reaching large state coincides with the task (a walker's forward position and velocity are themselves the large variables the task pays for), this crude bonus also improves on the baseline, on HalfCheetah by more than the epiplexity bonus does. But it is unstable: on Hopper, where large state means an extreme posture the agent cannot hold, the magnitude bonus drops far below the task reward, driving the agent to fall; on LunarLander it drops harder still, from a return of~$+169$ to~$-171$. The epiplexity bonus never collapses: across all ten tasks it stays at or above the task reward except on Walker2d, where it falls~$4\%$ short, while the magnitude control drops below the baseline on three tasks and far below on two. What sets learnable novelty apart is not a higher ceiling on any one task but its stability: it rewards state the bounded observer can compress, never the raw magnitude that aids one environment and ruins another.

The epiplexity-only column shows what learnable novelty alone can do. An agent rewarded only for its trajectory's epiplexity never sees the task, and on most of these tasks it does not perform it: optimizing learnable novelty is not optimizing the goal, and on Acrobot reaching the goal would end the episode and cut off the agent's own learnable novelty, so the epiplexity-only agent learns to avoid it. Learnable novelty alone is therefore not a task solver; its role is as a bonus: it supplies the exploration the task reward lacks while the task reward supplies the goal. The exceptions are Hopper and Walker2d, where moving forward is itself the richest trajectory: the epiplexity-only agent learns to locomote on its own, and on Walker2d it even exceeds the task reward itself (Table~\ref{tab:rl}). The bonus does not help everywhere: on tasks where the task reward already affords adequate exploration, the learnable-novelty drive is a mild distraction (Walker2d, Table~\ref{tab:rl}). But it helps exactly where exploration is the bottleneck, which is where an intrinsic drive is supposed to.

The prior work most directly related to these experiments is prediction-error curiosity~\citep{pathak2017curiosity, burda2019rnd}: it likewise points the reward at states an internal predictor has not yet learned, but that reward must be read off a model trained alongside the exploration. The reservoir estimator removes this inner training loop: the bounded observer is fixed, and what it can learn is computed in closed form. Prediction-error exploration can also be trapped by environmental noise, whereas the learnable-novelty reward inherits epiplexity's immunity to the noisy-television problem (Section~\ref{sec:novelty}): a region of pure noise inflates the agent's per-step surprise yet adds nothing to the learnable novelty, so the reward is not drawn to noise. What it rewards is movement into state the bounded observer can still compress, not state it merely cannot predict.

\section{Discussion}
\label{sec:discussion}

Complexity generation, abstraction, and exploration are ordinarily studied in separate fields and driven by objectives that owe nothing to one another. In the experiments reported here, all three were produced by a single quantity evaluated by a fixed observer of a single construction. Read as a measure, it recovered the classical complexity ordering of the elementary cellular automata without supervision, placing the one rule proven Turing-complete at the top of the space. Ascended as an objective, its gradient carried a neural cellular automaton from simple dynamics into a regime of complex solitons and organized the representation of an image encoder around the digit classes of MNIST, although no label ever entered training. Handed to an agent as an intrinsic reward, it supplied the exploration that task rewards lack, improving on the task reward in nine of ten environments and collapsing in none. The three phenomena, these results suggest, were never independent: they are projections of one quantity, learnable novelty, onto dynamics, representations, and behavior.

What separates this account from earlier theories of intelligence is that it places the observer at its center, inside the definition itself. All three experiments position a bounded observer of the same construction at the center of dynamical evolution, representation learning, and policy execution, and the quantity being maximized is defined relative to it. The intrinsic complexity of an automaton, the quality of a code, and the adaptivity of a policy thereby cease to be absolute properties of a system and become properties of a relationship: how much structure this particular bounded observer can extract from that system. The centrality of the observer has important precedents. Dennett's intentional stance treats belief and agency as attributes an interpreter ascribes rather than intrinsic facts~\citep{dennett1987intentional}; Pattee's epistemic cut locates the division between observer and observed within any act of measurement~\citep{pattee2001physics}; and the same commitment runs through second-order cybernetics and relational biology~\citep{vonfoerster1981observing, rosen1991life}. These traditions place the observer within the account of what can be said about a system. We carry this observer-centered view into a theory of intelligence, treating complexity generation, abstraction, and adaptive behavior as expressions of the relation between a system and a bounded learner. More recently, \citet{finzi2026epiplexity} formalize the structure available to a computationally bounded observer as epiplexity, the length of the program it distills from data. We read the same quantity as novelty: under prequential coding~\citep{dawid1984prequential}, the cumulative surprise of a learning observer splits into a learnable part and an unlearnable residual, and the two classical drives fail on opposite sides of this split: novelty search maximizes the sum and is dragged toward the residual, while the free-energy principle minimizes the sum and discards structure along with noise. Learnable novelty is the part both objectives needed and neither isolates.

Information theory has always contained an observer, but a silent one. Shannon's receiver has unbounded compute and therefore no character of its own; the observer presupposed by Kolmogorov complexity is likewise free of resource limits. Predictive information~\citep{bialek2001predictability}, the mutual information between the past of a stream and its future, comes closest to separating structure from randomness, yet still for that ideal observer: it counts whatever is predictable in principle, whether or not a bounded learner could extract it. Computational mechanics sharpens this into a construction, measuring extractable structure as the statistical complexity of the minimal set of causal states an observer needs to predict a process~\citep{crutchfield1989inferring}; but that observer, too, is idealized, charged nothing for the compute its reconstruction demands. Resource-bounded refinements gave the observer a budget, yet it remained a static parameter in the definition of a measure: something a quantity is stated relative to, not an agent in its own right. Our experiments argue for a more active role, and we propose an operational definition to match: the observer is a learner, a system that receives sequential data, pays the cost of its surprise, and updates its predictor with every payment. This view also agrees with the premise of AIXI~\citep{hutter2005universal}, where all of an agent's intelligence derives from the quality of the Solomonoff predictor~\citep{solomonoff1964formal}, an idealized sequential learner: the capacity to learn sets the ceiling on the capacity to behave intelligently.

The closest predecessor of the present objective is Schmidhuber's compression progress~\citep{schmidhuber2010curiosity, schmidhuber2009ultimate}. The two share their premise (learning is compression), but their optima part company. Compression progress rewards the rate at which the description of experience shrinks, and the total shrinkage available is set by how far that description stands above its Kolmogorov complexity. An utterly bland, unchanging stretch of experience has minimal Kolmogorov complexity and therefore the largest room to shrink, so in the long run a maximizer of cumulative progress profits most from trajectories that compress well because they contain little: this rebuilds the dark-room problem~\citep{sun2020darkroom}. Learnable novelty closes this route by construction: the program a bounded observer extracts from a dark room has length near zero, and no amount of further observation makes it grow.

Two limitations bound these results and point to a direction for future work. The first is that our observer never grows. Freezing the reservoir at initialization is what buys the closed form, but it also fixes the boundary of the learnable once and for all: structure beyond the reservoir's compressive reach is invisible to the score, and once an optimized system exhausts what its observer can read, learnable novelty saturates. In its present form the estimator captures only the shallow structure that is linearly readable from random features. The natural way past this is to let the observer into the optimization. When observer and observed co-evolve, the system is pushed to produce whatever structure its observer can currently absorb while the observer extends its reach in step, and the boundary of the learnable moves rather than the score saturating against it. Open-ended growth, on this account, asks for exactly that moving boundary. Large language models may be a ready substrate for this symmetric arrangement: through in-context learning a model behaves as a compute-bounded sequential learner, and as a generator of sequences it is at the same time the system being observed.

The second limitation concerns what learnable novelty is to the agent that pursues it. In our reinforcement-learning experiments it enters as a bonus: it reliably relieves the exploration bottleneck, but the direction of optimization is still anchored by a task reward imposed from outside. One extension would invert the relationship, converting the task reward into a modulation of where novelty is to be found, so that the agent's underlying drive remains the pursuit of learnable novelty and the environment serves only to shape the landscape through which that pursuit moves. The environments tested here are also, almost without exception, finite games, in which the surest way to keep novelty flowing is to avoid terminal states; an agent rewarded for learnable novelty alone accordingly learns to survive, and sometimes to defer the very goal that would end the episode. Many of the settings that matter most are infinite games with no terminal state to guard against, and there the condition this framework adds to the causal-entropic account of intelligent behavior~\citep{wissner2013causal} should carry more of the weight.

Compression, computation, and exploration need not be assigned separate, mutually isolated objectives: on three very different substrates, each emerged from the sustained pursuit of learnable novelty by a bounded observer. Taking the bounded learner as the primitive, restating existing theories from its standpoint, and letting observer and observed grow together are, we believe, key questions for future exploration.

\bibliography{ref}
\bibliographystyle{plainnat}

\appendix

\clearpage

\raggedbottom

\section{Architectures and sampling}
\label{app:architectures}

\subsection{Elementary cellular automata}

There are~$256$ elementary rules in total; after removing those equivalent under the left--right reflection and the~$0 \leftrightarrow 1$ colour swap, $88$ locally unique rules remain (the constant rule~0 among them), and these are the ones scored in the main text. Each elementary rule~$r$ has a binary update map~$F_r$ acting on a width-$64$ line with circular boundary conditions. For each rule we draw~$512$ independent initial states~$x_i^{(0)} \sim \mathrm{Bernoulli}(1/2)$, evolve each for a~$1000$-step burn-in so that it lands on the rule's attractor, and take as target the stacked window of its next~$32$ one-step states,
\[
    x_i = F_r^{1000}\big(x_i^{(0)}\big), \qquad Y_i = \big(F_r(x_i),\, F_r^{2}(x_i),\, \dots,\, F_r^{32}(x_i)\big),
\]
where~$F_r^{k}$ denotes~$k$ applications of the rule. We repeat the scoring ten times per rule, each time redrawing both the random reservoir weights and the input ensemble, and report the mean and standard deviation of the resulting epiplexity. Table~\ref{tab:eca} lists these settings together with the reservoir used to score them.

\begin{table}[tb]
\centering
\caption{Sampling and reservoir settings for the elementary cellular automata.}
\label{tab:eca}
\begin{tabular}{ll@{\hspace{2.5em}}ll}
\toprule
\multicolumn{2}{l}{\textit{Sampling}} & \multicolumn{2}{l}{\textit{Reservoir}} \\
\midrule
Lattice width &~$64$ & Feature map & circular 1D convolution \\
Boundary conditions & circular & Depth &~$3$ \\
Initialization & Bernoulli$(1/2)$ & Channels &~$256$ \\
Burn-in steps &~$1000$ & Kernel size &~$3$ ($1$ in the final layer) \\
Stacked target window &~$\tau=32$ & Activation & ELU \\
Samples per rule~$N$ &~$512$ & Ridge~$\lambda$ &~$0.03$ \\
\bottomrule
\end{tabular}
\end{table}

\subsection{Reservoir feature maps}

The bounded observer is a fixed, randomly initialized reservoir whose architecture matches the data geometry. Table~\ref{tab:reservoirs} lists the configurations used in this paper.

\begin{table}[tb]
\centering
\caption{Reservoir feature maps used as bounded observers. Every map places a pre-activation normalization before each nonlinearity, holding it at the edge of chaos independent of depth and architecture (Appendix~\ref{app:criticality}).}
\label{tab:reservoirs}
\begin{tabular}{ll p{6.2cm}}
\toprule
Data geometry & Feature map~$\phi$ & Settings \\
\midrule
1D cellular automata & circular 1D convolution & $256$ channels, kernel~$3$ ($1$ in the final layer), ELU; depth~$3$, $\lambda = 0.03$ (ECA ranking); depth~$4$, $\lambda = 0.3$ (inverse NCA) \\
Continuous-time flows & random MLP & depth~$4$, width~$64$, ELU, $\lambda = 0.1$ \\
Flattened images (MNIST encoder) & random MLP & depth~$4$, width~$2048$, ELU, $\lambda = 3$, $\eta = 30$ \\
RL observation trajectories & random MLP & depth~$4$, width~$32$, ELU, $\lambda = 0.3$ \\
\bottomrule
\end{tabular}
\end{table}

\subsection{Reservoir criticality: a pre-activation normalization at the edge of chaos}
\label{app:criticality}

The reservoir is random and untrained, so unlike a learned network nothing during fitting moves it toward a useful representation: whether its nonlinearity is engaged at all is decided by its initialization and by the scale of its input. A plain random feedforward stack is a poor default here. Driven by an~$O(1)$ input, the pre-activations of a depth-four ELU reservoir have a standard deviation of only a few tenths, so each ELU operates near the origin where it is indistinguishable from a linear map, and the feature map it computes is close to a random linear projection. The operating point is also unstable: it drifts with depth and shifts by orders of magnitude with the input scale, so one architecture can be nearly linear in one experiment and saturated in another.

For recurrent reservoirs the analogous question is settled by the echo-state property and the spectral radius. The underlying principle is that a reservoir is most expressive at the \emph{edge of chaos}, the boundary between an ordered regime where signals contract and a chaotic one where they explode~\citep{langton1990computation, packard1988adaptation}. We define the order parameter as the per-layer perturbation multiplier
\begin{equation}
    \chi = \E\,\frac{\lVert \delta h^{(\ell+1)} \rVert}{\lVert \delta h^{(\ell)} \rVert},
\end{equation}
the factor by which an infinitesimal input perturbation grows from one layer to the next~\citep{poole2016exponential, schoenholz2017deep}:~$\chi < 1$ is the ordered phase, where signal and gradients vanish with depth,~$\chi > 1$ the chaotic phase, and~$\chi \approx 1$ the edge of chaos. Measured by finite differences, the plain reservoir sits in the ordered phase for every architecture we test, at~$\chi \approx 0.55$ for depths up to sixteen and rising to~$\chi \approx 0.75$ at depth thirty-two, always well below criticality.

\begin{figure}[tb]
    \centering
    \includegraphics[width=\textwidth]{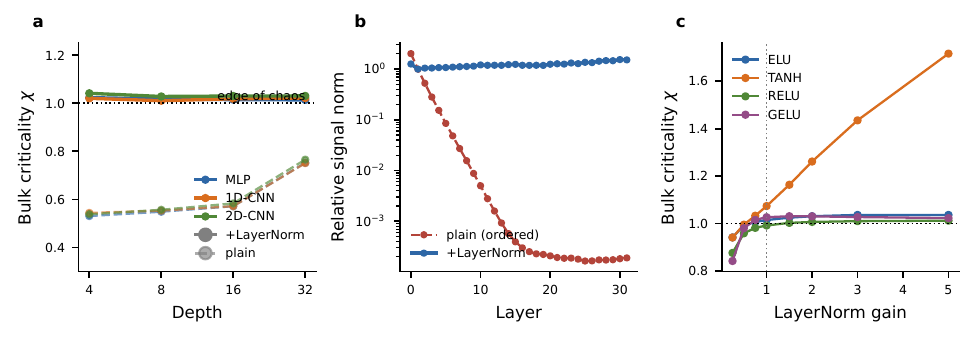}
    \caption{A pre-activation normalization holds a random reservoir at the edge of chaos. \textbf{(a)}~Bulk criticality~$\chi$ against depth for MLP, 1D-, and 2D-convolutional reservoirs: the plain reservoir (dashed) sits in the ordered phase, well below one, while the normalized one (solid) holds at~$\chi \approx 1$ for every architecture and depth. \textbf{(b)}~Signal survival through depth, the relative norm of an input perturbation by layer: the plain reservoir contracts it exponentially, the normalized one preserves it. \textbf{(c)}~Bulk~$\chi$ against the normalization gain for four activations; gain one reaches~$\chi \approx 1$ for all of them.}
    \label{fig:criticality}
\end{figure}

We find that a single rule suffices to hold the reservoir at the edge of chaos: normalize the pre-activations over the feature (channel) axis immediately before every nonlinearity. Fixing the pre-activation scale to unit variance places each nonlinearity in its responsive region and pins~$\chi$ to a value the activation alone sets, independent of depth, input scale, and architecture (Figure~\ref{fig:criticality}).

\subsection{Trainable systems}

The neural cellular automaton, the MNIST encoder, and the reinforcement-learning policy are the systems trained to maximize learnable novelty. Table~\ref{tab:trainable} lists their architectures and optimization settings; the reservoirs they are scored against are in Table~\ref{tab:reservoirs}.

\begin{table}[H]
\centering
\caption{Trainable systems and their optimization settings.}
\label{tab:trainable}
\begin{tabular}{l p{9.8cm}}
\toprule
\multicolumn{2}{l}{\textit{Neural cellular automaton} ($x_{t+1} = \operatorname{normalize}(x_t + g_\theta(x_t))$, circular boundary)} \\
\midrule
Lattice width &~$64$ \\
Lift convolution & radius~$2$ (kernel~$5$) to~$128$ channels, batch normalization, GELU \\
Hidden layer & one pointwise convolution with batch normalization and GELU \\
Output projection & pointwise convolution to the two state channels \\
Burn-in steps &~$32$ (no gradient tracking) \\
State noise & Gaussian, standard deviation~$0.1$, added after burn-in \\
Stacked target window &~$\tau=8$ \\
Batch size &~$2048$ \\
State constraint & two-channel field, unit norm per site \\
Optimizer & AdamW with cosine annealing, learning rate~$1 \times 10^{-4}$, $S^\phi$ scaled by~$1/100$ before the gradient step \\
Gradient clipping &~$0.5$ \\
Training steps &~$2{,}000$ \\
\midrule
\multicolumn{2}{l}{\textit{MNIST encoder} (code~$z = E_\theta(x)$, no labels)} \\
\midrule
Encoder~$E_\theta$ & trainable MLP with hidden widths $64, 128, 256$ and code dimension~$D = 64$ \\
Code constraint & unit norm, $z / \lVert z \rVert$ \\
Batch size &~$128$ \\
Optimizer & AdamW with cosine annealing \\
Training steps &~$500$ \\
\midrule
\multicolumn{2}{l}{\textit{Reinforcement-learning policy} (PPO, default MLP policy)} \\
\midrule
Algorithm & PPO~\citep{schulman2017ppo, raffin2021stable} \\
Parallel environments &~$8$ \\
Rollout length &~$1024$ \\
Minibatch &~$256$ \\
Discount~$\gamma$ &~$0.999$ \\
GAE~$\lambda_{\mathrm{GAE}}$ &~$0.98$ \\
Learning rate &~$3 \times 10^{-4}$ \\
Training steps &~$600{,}000$ \\
\bottomrule
\end{tabular}
\end{table}

\begin{figure}[!t]
    \centering
    \includegraphics[width=\textwidth]{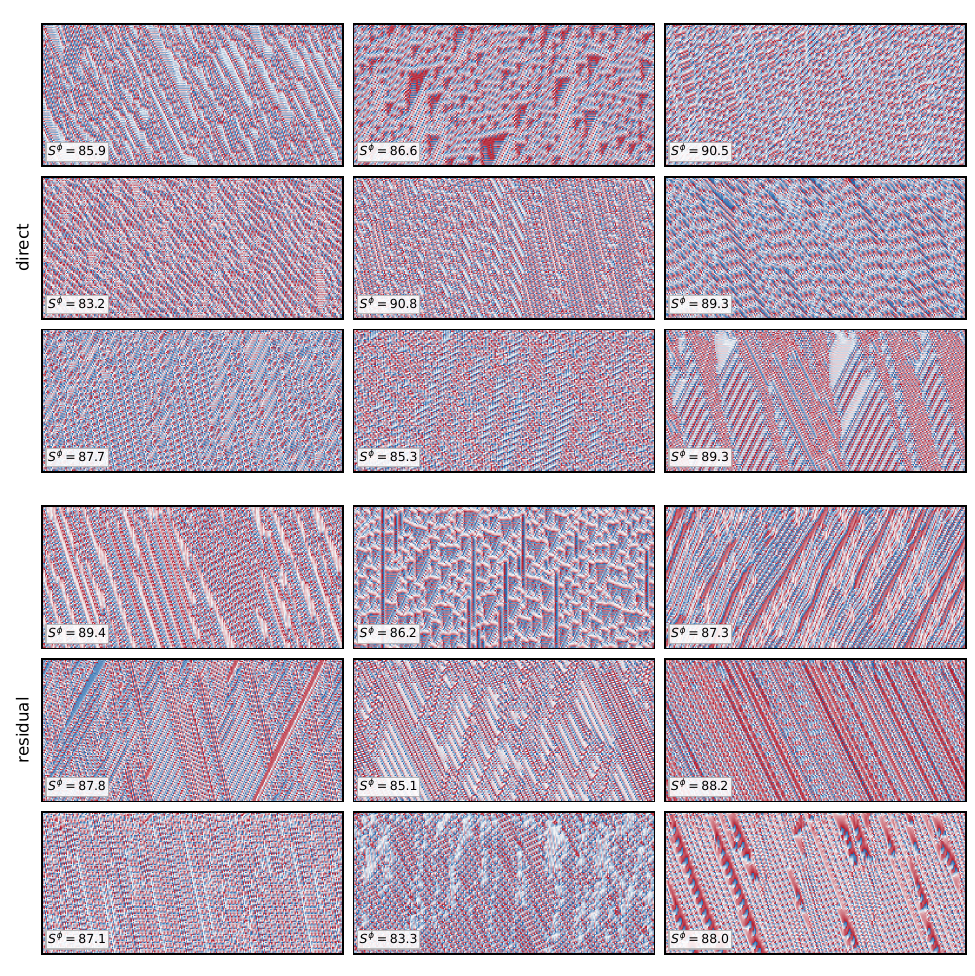}
    \caption{Final NCA evolution under the direct and residual update rules, each over nine random seeds, with the architecture and target window held fixed. The upper three rows use the direct normalized update, the lower three the residual update; the nine cells of each block are nine independent seeds. Every panel shows a final-checkpoint rollout and annotates its final~$S^\phi$. Both rules develop coherent traveling structures across all nine seeds, with final scores in a narrow band.}
    \label{fig:nca-variants}
\end{figure}

For the neural cellular automaton, \(g_\theta\) denotes the convolutional stack in Table~\ref{tab:trainable}, and \(\operatorname{normalize}\) divides the two-channel vector at each lattice site by its Euclidean length, \(\operatorname{normalize}(v)_i = v_i / \lVert v_i \rVert\) over the two channels at site~\(i\), so every site lies on the unit circle. The main-text run uses the residual normalized update, which keeps a skip connection before the unit-norm projection,
\[
    x_{t+1} = \operatorname{normalize}\big(x_t + g_\theta(x_t)\big).
\]
A direct variant drops the skip connection and normalizes the convolutional output alone,
\[
    x_{t+1} = \operatorname{normalize}\big(g_\theta(x_t)\big).
\]
For both update forms, training samples unit-norm two-channel initial states on a width-\(64\) ring, applies \(32\) no-gradient burn-in steps, and adds independent Gaussian noise of standard deviation \(0.1\) at every site of the burned-in state; the stacked future window \(Y=(G_\theta(X), \dots, G_\theta^{\tau}(X))\) is rolled out and scored from this perturbed state. Without the perturbation, a run that settles onto a spatially uniform fixed point presents the reservoir with a constant input--target pair, whose score is zero and provides no gradient to escape it. Figure~\ref{fig:nca-variants} holds the architecture, optimizer, and target window fixed and contrasts the two update rules, training each from nine independent random seeds. Both rules develop coherent traveling structures across all nine seeds.

\section{Robustness of the elementary-cellular-automaton ranking}
\label{app:eca-robustness}

The ranking of Figure~\ref{fig:eca-combined} is computed under the fixed estimator configuration of Table~\ref{tab:eca}. To test how much of it depends on that choice, we vary one hyperparameter at a time around the reference configuration and re-score all $88$ rules: the resolution~$\eta \in [0.03, 30]$, the ridge penalty~$\lambda \in [0.001, 1]$, the target window length~$\tau \in \{4, \dots, 64\}$, the reservoir depth~$\in \{1, \dots, 5\}$, kernel size~$\in \{3, 5, 7\}$, channel count~$\in \{64, \dots, 512\}$, and the sample count~$N \in \{128, \dots, 1024\}$. Each configuration is scored with a single draw of the reservoir weights and input ensemble; the reference configuration, every configuration at which the top rank changes, and every configuration at which the score order of rules~30 and~54 flips are re-scored with three independent draws.

\begin{figure}[tb]
    \centering
    \includegraphics[width=\textwidth]{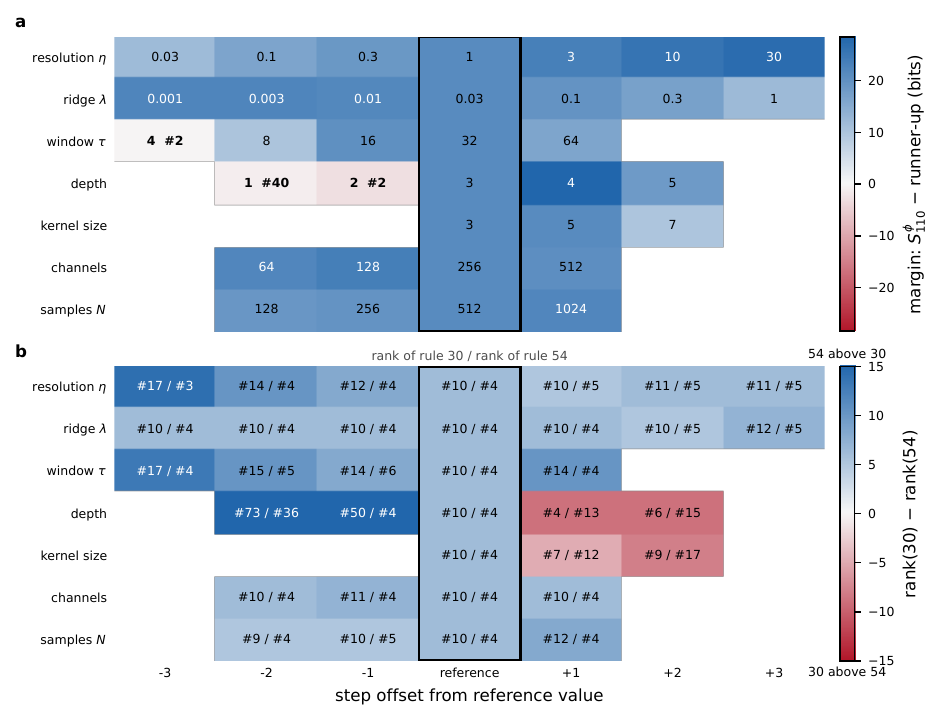}
    \caption{One-at-a-time robustness of the ECA ranking. In both panels each row varies one estimator hyperparameter around the reference configuration of Table~\ref{tab:eca} (outlined column), with columns aligned by step offset from the reference value. \textbf{(a)}~Margin, in bits, between rule 110's epiplexity and that of the best other rule; each cell prints the value scored, so a positive (blue) cell is a configuration at which rule 110 ranks first, and in the cells where it is not first the cell adds rule 110's rank in bold (single draw). \textbf{(b)}~Ranks of the reference rules 30 (chaotic, before the slash) and 54 (complex, after the slash) under the same scan; color gives their difference, blue where rule 54 ranks above rule 30 and red where rule 30 ranks above rule 54, saturated at $\pm 15$.}
    \label{fig:eca-robustness}
\end{figure}

Rule~110 keeps the top rank across the entire tested range of every axis except the two that cut the observer or the target window below the scale of its structure (Figure~\ref{fig:eca-robustness}a). At the reference configuration its margin over the best other rule (rule~25) is $20.7$ bits over three draws, and the margin stays between $9.6$ and $28.4$ bits at every other configuration where it is first. The full ranking is similarly rigid: its Spearman correlation with the reference ranking stays above $0.90$ everywhere except at depth~$\leq 2$. The first exception is $\tau = 4$, where the window is too short for the accumulated structure of rule~110 to separate it from the field and it falls $0.5$ bits behind rule~73 ($18.2 \pm 0.4$ against $18.7 \pm 0.2$, three-draw means). The second is small depth: at depth~$2$ the reservoir's receptive field has radius one and rule~25 overtakes rule~110 ($13.5 \pm 0.4$ against $10.4 \pm 0.5$ bits), and at depth~$1$ the reservoir sees a single cell and the measurement collapses altogether (rule~110 falls to rank~$40$ and the correlation with the reference ranking vanishes).

The reference rules~30 and~54 order themselves by the observer's locality (Figure~\ref{fig:eca-robustness}b). At the reference configuration the complex rule~54 ranks fourth and the chaotic rule~30 twelfth, with a score gap $S^\phi_{54} - S^\phi_{30} = 10.9 \pm 1.0$ bits over three draws, and this order holds across the full tested range of the resolution, ridge, window, channel, and sample axes. It reverses exactly where the receptive field grows beyond radius two: one more layer (depth~$4$, radius three) puts rule~30 fifth and rule~54 thirteenth (gap $-14.6 \pm 1.0$ bits), and a wider kernel (kernel~$5$, radius four) does the same ($-11.9 \pm 0.5$ bits). An observer of radius at most two prices the structured rule above the chaotic one; a wider view makes the chaotic rule's short-range structure learnable and reverses the order. Rule~110 ranks first in both regimes.

\section{The log-determinant description length as a matrix-\texorpdfstring{$t$}{t} marginal}
\label{app:matrix-t}

The main text prices the readout directly by the spectral description length~\eqref{eq:spectral-code}. The same form arises from a hierarchical Gaussian prior. Let~$W \in \R^{m \times D}$ and make its conditional law matrix Gaussian,
\begin{equation}
    W \mid \Lambda \;\sim\; \gM\gN_{m, D}\big(0, \Lambda^{-1}, I_D\big),
    \qquad
    p(W \mid \Lambda) \;\propto\; |\Lambda|^{D/2} \exp\!\Big(-\tfrac{1}{2}\operatorname{tr}\big(\Lambda W W^\top\big)\Big),
\end{equation}
with~$\Lambda \succ 0$ a row precision matrix. Holding the precision fixed at~$\Lambda = \lambda I_m$ collapses the negative log-density, in bits, to a quadratic,
\begin{equation}
    -\log_2 p(W \mid \lambda I_m) \;=\; C + \frac{\lambda}{2 \ln 2}\,\|W\|_F^2 ,
\end{equation}
the ridge penalty. If instead the precision is not specified in advance but given an isotropic Wishart-type hyperprior~$p(\Lambda) \propto |\Lambda|^{(a - m - 1)/2} \exp\!\big(-\tfrac{1}{2\eta}\operatorname{tr}\Lambda\big)$, with~$a > m - 1$ so that the prior is proper and the integrals below converge, the marginal density of~$W$ is~$p(W) = \int_{\Lambda \succ 0} p(W \mid \Lambda)\, p(\Lambda)\, d\Lambda$. Collecting the powers and exponential terms in~$\Lambda$,
\begin{equation}
    p(W) \;\propto\; \int_{\Lambda \succ 0} |\Lambda|^{(a + D - m - 1)/2} \exp\!\Big[-\tfrac{1}{2}\operatorname{tr}\big((W W^\top + \eta^{-1} I_m)\Lambda\big)\Big]\, d\Lambda ,
\end{equation}
and applying the Wishart integral identity~$\int_{\Lambda \succ 0} |\Lambda|^{(\nu - m - 1)/2} \exp\!\big(-\tfrac{1}{2}\operatorname{tr}(A\Lambda)\big)\, d\Lambda \propto |A|^{-\nu/2}$ with~$\nu = a + D$ and~$A = W W^\top + \eta^{-1} I_m$ gives
\begin{equation}
    p(W) \;\propto\; \det\!\big(W W^\top + \eta^{-1} I_m\big)^{-(a + D)/2} \;\propto\; \det\!\big(I_m + \eta\, W W^\top\big)^{-(a + D)/2} .
\end{equation}
Dropping the constants independent of~$W$,
\begin{equation}
    -\log_2 p(W) \;=\; C + \frac{a + D}{2}\,\log_2 \det\!\big(I_m + \eta\, W W^\top\big) ,
\end{equation}
which is the spectral description length~\eqref{eq:spectral-code} with~$\alpha = (a + D)/2$. The log-determinant form is therefore the description length of the Gaussian readout after the unknown precision has been integrated out: the ridge penalty corresponds to a fixed precision, the log-determinant to a marginalized one. The marginalization supplies the \emph{form} of the description length; the estimator treats~$\alpha$ as a free overall scale and fixes~$\alpha = 1/2$ throughout, rather than the~$D$-dependent value the hierarchical derivation would assign.

\section{Exact minimization of the description length}
\label{app:ridge-surrogate}

The main text fits the readout by a ridge approximation to the total description length (Section~\ref{sec:method}). This appendix minimizes the objective directly, without the approximation:
\begin{equation}
    \gJ_{\mathrm{MDL}}(W)
    \;=\;
    \frac{1}{2\sigma^2 \ln 2}\,\|\tilde{Y} - \tilde{H} W\|_F^2
    \;+\;
    \gC_{\mathrm{spec}}(W),
    \label{eq:mdl-objective}
\end{equation}
with~$\gC_{\mathrm{spec}}$ the spectral description length~\eqref{eq:spectral-code} and the residual scale calibrated as~$\sigma^2 = \lambda / (2\alpha\eta)$, so that the two objectives coincide in the small-readout regime, where~$\log\det(I_m + \eta\, W W^\top) \approx \eta\,\|W\|_F^2$ and the spectral description length reduces to the ridge penalty. We write~$S^\phi_{\mathrm{MDL}}$ for the spectral description length of the exact optimum~$W_{\mathrm{MDL}}$.

The objective has no closed-form minimizer, but the log-determinant is concave in~$W W^\top$, so its linearization at the current iterate is a global upper bound on the description length, and minimizing that bound is a weighted ridge solve:
\begin{equation}
    W_{k+1}
    \;=\;
    \big(\tilde{H}^\top \tilde{H} + \lambda\, M_k\big)^{-1} \tilde{H}^\top \tilde{Y},
    \qquad
    M_k \;=\; \big(I_m + \eta\, W_k W_k^\top\big)^{-1}.
    \label{eq:mdl-mm}
\end{equation}
Each sweep of this majorize--minimize iteration decreases~$\gJ_{\mathrm{MDL}}$ monotonically. Warm-started at~$W_\lambda$ and run in double precision on the sufficient statistics, it converges to a stationary point; rescaling the warm start by factors between~$1/4$ and~$4$ leaves the solution unchanged.

\begin{figure}[tb]
    \centering
    \includegraphics[width=\textwidth]{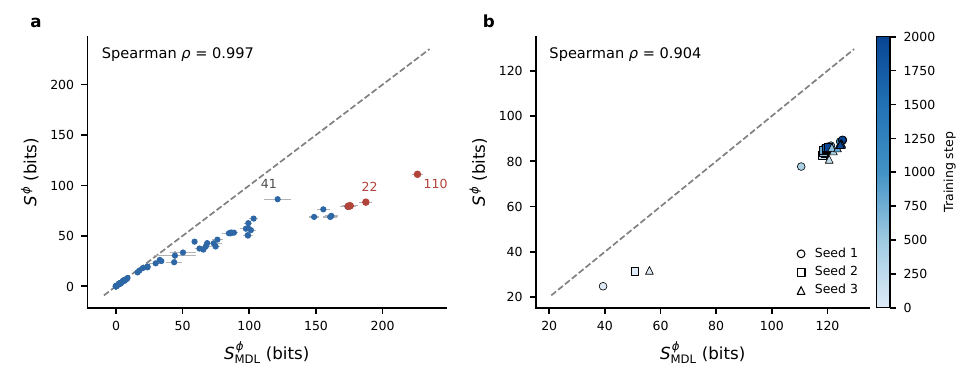}
    \caption{The ridge readout against the exact minimizer of the description length, on identical reservoirs and data. \textbf{(a)}~Per-rule mean scores for the~$88$ elementary rules: the reported~$S^\phi$ (the spectral description length of the ridge readout~$W_\lambda$) against~$S^\phi_{\mathrm{MDL}}$ (that of the minimizer of~\eqref{eq:mdl-objective}). Error bars are standard deviations over ten reservoir and data draws; the dashed line is the identity; red marks the five highest rules under~$S^\phi_{\mathrm{MDL}}$, and rule~41 is the largest rank change (second to thirteenth). \textbf{(b)}~The same comparison at every checkpoint of the three inverse-NCA training runs (marker shape: seed; colour: training step; error bars: standard deviations over three evaluation draws).}
    \label{fig:surrogate-vs-mdl}
\end{figure}

Figure~\ref{fig:surrogate-vs-mdl} compares the two readouts on identical reservoirs and data. The exact minimizer extracts more bits from the same data: across the~$88$ elementary rules (ten reservoir and data draws each, the setting of Table~\ref{tab:eca}), $S^\phi_{\mathrm{MDL}}$ exceeds~$S^\phi$ by a factor growing from about~$1$ to about~$2.3$ with the score itself, because the log-determinant description length is flatter than the quadratic penalty at large singular values and lets the readout reach further into faintly expressed directions. The excess is one-sided: under an isotropic feature Gram every stationary point of~\eqref{eq:mdl-objective} satisfies~$w\,\big(1 + \lambda/(1 + \eta w^2)\big) = \sigma$ along each singular direction and so shrinks less than the ridge solution~$\sigma/(1+\lambda)$, giving~$S^\phi \le S^\phi_{\mathrm{MDL}}$; the measured Gram is far from isotropic, but the ordering holds in every paired solve we ran ($880$ for the ECA, $99$ for the NCA). Yet the two readouts rank systems almost identically: Spearman~$\rho = 0.997$ over the rules, the same top-fourteen set, rule~110 maximal under both with its margin over the runner-up widening from~$24.7$ to~$38.7$ bits, and the largest rank change anywhere is rule~41, second under the ridge readout and thirteenth under the exact one; on the inverse-NCA trajectories (three seeds, eleven checkpoints each) the two scores move together (Pearson~$r = 0.996$), with a nearly constant~$35$-bit offset on the converged plateau. Removing the approximation changes how many bits the observer extracts, not which systems it finds structured; the estimator therefore keeps the ridge readout.

\section{Numerically stable, differentiable ridge solve}
\label{app:ridge-solve}

The estimator in~\eqref{eq:rc-score} is defined through the ridge optimum~$W_\lambda = (\tilde{H}^\top \tilde{H} + \lambda I_m)^{-1} \tilde{H}^\top \tilde{Y}$. Solving it the textbook way, by forming the Gram matrix~$\tilde{H}^\top \tilde{H}$ and inverting (the normal equations), squares the conditioning of the design matrix, turning a condition number~$\kappa(\tilde{H})$ into~$\kappa(\tilde{H})^2$. In single precision this is not a minor loss: whenever the reservoir features are nearly collinear, which is exactly the low-epiplexity regime where the bounded readout finds little independent structure to use, the squared conditioning corrupts~$W_\lambda$ and therefore the score. Because the inverse experiments drive the generating system across a wide range of feature conditioning, an unstable solve distorts both~$S^\phi$ and the gradient the optimization follows.

The batch solve therefore never forms the Gram matrix. For output coordinate~$j$, writing the ridge objective as a single least-squares problem gives
\begin{equation}
    \big\|\tilde{H} w_j - \tilde{y}_j\big\|_2^2 + \lambda \|w_j\|_2^2
    \;=\;
    \left\|
    \begin{bmatrix} \tilde{H} \\ \sqrt{\lambda}\, I_m \end{bmatrix} w_j
    -
    \begin{bmatrix} \tilde{y}_j \\ 0_m \end{bmatrix}
    \right\|_2^2 ,
    \label{eq:augmented-ls}
\end{equation}
where~$\tilde{y}_j$ is column~$j$ of~$\tilde{Y}$. The ridge optimum is the ordinary least-squares solution of the augmented system with design matrix~$\tilde{H}_{\mathrm{aug}} = [\,\tilde{H};\ \sqrt{\lambda}\, I_m\,] \in \R^{(N + m) \times m}$ and target~$\tilde{y}_{j,\mathrm{aug}} = [\,\tilde{y}_j;\ 0_m\,]$. We solve this with a reduced QR factorization~$\tilde{H}_{\mathrm{aug}} = QR$ followed by a triangular solve,
\begin{equation}
    w_j = R^{-1} Q^\top \tilde{y}_{j,\mathrm{aug}},
    \label{eq:qr-solve}
\end{equation}
and stacking the column solutions gives~$W_\lambda = [\,w_1,\dots,w_D\,]$. This operates on the design matrix directly. The effective conditioning is that of~$\tilde{H}_{\mathrm{aug}}$, the square root of what the normal equations would face, and the whole computation stays in the input dtype without ever materializing~$\tilde{H}^\top \tilde{H}$. The~$\sqrt{\lambda}\, I_m$ block gives~$\tilde{H}_{\mathrm{aug}}$ full column rank for any~$\lambda > 0$, so~$R$ is invertible and the solve is well posed even when~$\tilde{H}$ is itself rank-deficient.

This solve is differentiable. Both the QR factorization and the triangular solve have well-defined derivatives, so~$W_\lambda$ (and hence the spectral score~$S^\phi = \alpha \log_2 \det(I_m + \eta\, W_\lambda W_\lambda^\top)$, whose singular-value decomposition is likewise differentiable) is a differentiable function of~$\tilde{H}$ and~$\tilde{Y}$, and through the normalization and the frozen reservoir~$\phi$, of the data~$(X, Y)$. In the inverse experiments we backpropagate~$S^\phi$ through exactly this computation into the parameters of the system that generates~$(X, Y)$, with~$\phi$ held fixed. The closed-form solve therefore serves both as a cheap forward score and as a differentiable layer in the optimization.

\section{Reservoir epiplexity on continuous-time flows}
\label{app:observer-extra}

The same closed-form score, with its reservoir an MLP on the state vector, ranks a further family of dynamical systems, continuous-time flows, consistently with a longstanding qualitative classification.

\subsection{Continuous-time chaotic systems}

For continuous-time flows we use the same random MLP reservoir as the other feedforward experiments: a four-layer network with ELU activations, hidden width~$64$, and ridge penalty~$\lambda = 0.1$. Each sample pairs an attractor state~$x$ with a stacked window of its future: we draw a random initial condition, integrate the system for a burn-in time of~$0.5$ to reach the attractor, and record the state as~$x$; R\"ossler and Thomas receive a longer initial integration before this burn-in (a pre-burn-in), because their attractors lie far from the region where initial conditions are drawn and the $0.5$-unit burn-in alone would not reach them; then we snapshot the trajectory at the ten lead times~$0.1, 0.2, \dots, 1.0$ from~$x$ and stack them as the target window~$y$. Both~$x$ and every target snapshot are standardized per state dimension by the empirical mean and standard deviation of the ensemble of~$x$, so the attractor's absolute scale does not enter the score. The reservoir maps~$x$ to features and one ridge readout per snapshot predicts the whole window; the epiplexity is the program length of the stacked readout, scored as in~\eqref{eq:rc-score}. We draw~$512$ samples per system and average the score over eight independent draws of the reservoir weights.

\begin{figure}[tb]
    \centering
    \includegraphics[width=\textwidth]{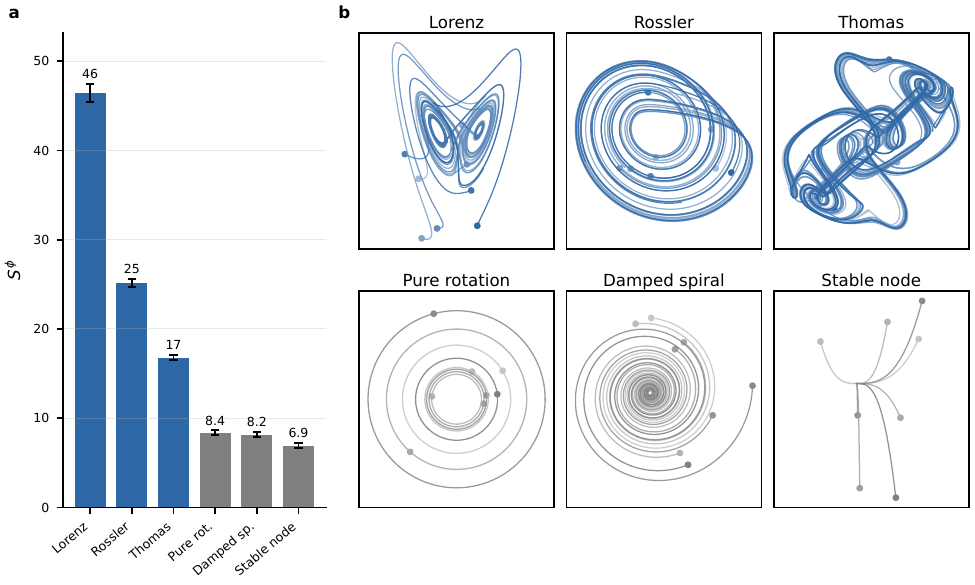}
    \caption{MLP-reservoir epiplexity on continuous-time systems. \textbf{(a)}~The chaotic systems (Lorenz, R\"ossler, Thomas) score well above the linear baselines. \textbf{(b)}~Phase portraits of the six systems, eight initial conditions each.}
    \label{fig:dynamical-systems}
\end{figure}

Across three chaotic systems (Lorenz, R\"ossler, Thomas) and three two-dimensional linear systems (pure rotation, damped spiral, stable node), the chaotic systems score well above the linear baselines (Figure~\ref{fig:dynamical-systems}). Lorenz scores highest at~$46$, with R\"ossler at~$25$ and Thomas at~$17$; all three sit well above the linear systems, which fall between~$6.9$ and~$8.4$.

\subsection{Temporal accumulation of epiplexity}
\label{app:temporal}

The cross-system comparison above draws independent state pairs from an ensemble of initial conditions. In the reinforcement-learning setting, where a dense per-step reward is needed, we instead track how epiplexity accumulates along a single trajectory as observations arrive one by one. Given a trajectory~$x_0, x_1, \dots, x_N$ sampled at interval~$\Delta t_s$, define the cumulative epiplexity~$S^\phi_t$ as the score computed from all pairs~$(x_i, y_i)$ with~$i \le t - \tau$, where~$y_i$ is the future window stacked from step~$i$ and~$\tau$ its horizon in samples. Each new observation adds one pair to the ridge regression; the marginal contribution~$\xi_t = S^\phi_t - S^\phi_{t-1}$ measures how much learnable structure the latest state brings.

Recomputing the ridge readout from scratch at every step costs~$O(N^2 m^2)$ for a length-$N$ stream. The same quantity admits an exact recursive least-squares form: maintaining the inverse Gram matrix and the readout under a Sherman--Morrison rank-1 update advances each step in~$O(m^2)$, plus~$O(m^2 D)$ for the singular values of the~$m \times D$ readout whenever the score itself is evaluated. Unlike the batch QR solve of Appendix~\ref{app:ridge-solve}, this covariance form does maintain the inverse Gram matrix and so gives up that solve's conditioning advantage; at the small reservoir widths used online ($m = 32$) this is numerically benign.

\section{Additional details for the inverse experiments}
\label{app:inverse-extra}

\subsection{Neural cellular automaton}

With the architecture and optimization settings of Table~\ref{tab:trainable}, the score climbs from its low initial value over the first several hundred steps, then rises more gradually and levels onto a plateau by about step~$1{,}500$, ending between~$S^\phi = 86$ and~$89$ across the three seeds of the residual~$\tau=8$ run shown in Figure~\ref{fig:nca-snapshots}. The traveling structures emerge early in the climb and, once formed, persist through the remainder of training. Rolling the final checkpoint out from a fresh random initial state on a wider lattice preserves the same regime beyond the training width, indicating that the learned local rule has not simply memorized the training-time lattice size.

\subsection{MNIST encoder}

The ridge regularizer~$\lambda$ controls how tight the readout bound is and therefore which encoders qualify as high-epiplexity. We use~$\lambda = 3$, far above the values typical of reservoir computing ($10^{-5}$ or smaller): a tight bound is what keeps redundancy expensive and forces the encoder toward compressed codes organized by the data's underlying factors, rather than letting it satisfy a loose readout with arbitrary ones. The raw score grows with both the code dimension~$D$ and the reservoir width, but a higher raw score reached by enlarging either does not by itself sharpen the class structure: it lets the encoder satisfy the readout without compressing. We report~$D = 64$ with a width-$2048$ reservoir; tightening the bound, rather than enlarging the model, is what selects for natural abstractions.

\begin{figure}[tb]
    \centering
    \includegraphics[width=\textwidth]{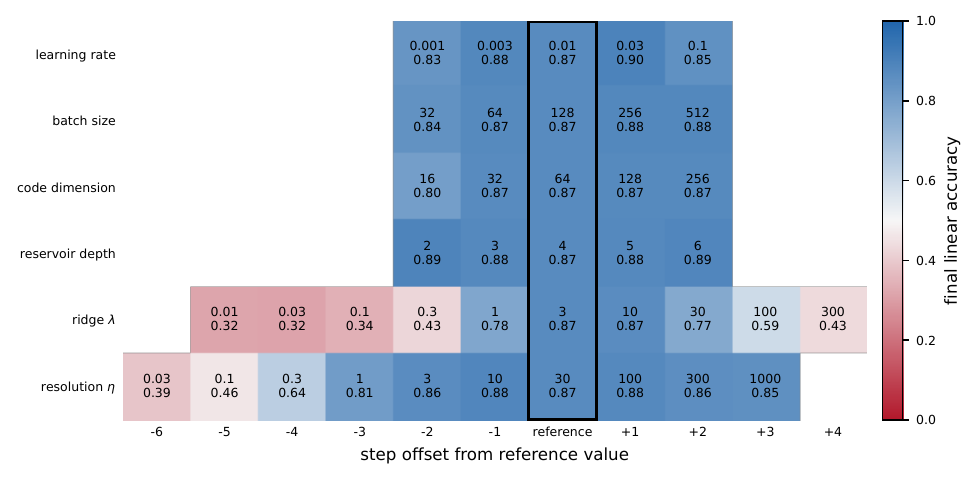}
    \caption{One-at-a-time robustness of the MNIST representation experiment. Each cell gives the varied hyperparameter (top) and final linear-probe accuracy after~$150$ training steps (bottom); all other settings remain at the reference values in the black-outlined column. Columns order the tested values within each row by their offset from the reference. The color scale is fixed from~$0$ to~$1$, centered at white for~$0.5$ accuracy; red marks values below~$0.5$ and blue values above it.}
    \label{fig:mnist-hparam-robustness}
\end{figure}

We test local robustness by varying one hyperparameter at a time around the reported configuration and training each variant for~$150$ steps with the same seed (Figure~\ref{fig:mnist-hparam-robustness}). Learning rate, batch size, code dimension, and reservoir depth leave the final linear-probe accuracy between~$0.80$ and~$0.90$ across their tested ranges. The estimator parameters have a wider failure regime: very weak ridge regularization ($\lambda \leq 0.3$) or low resolution ($\eta \leq 0.1$) drives accuracy below~$0.5$, whereas the neighborhood containing the reported $\lambda=3$ and $\eta=30$ remains high-accuracy. Thus the result does not depend narrowly on the precise reference values, although the estimator must still impose a sufficiently selective readout.

\section{Reinforcement-learning details}
\label{app:rl}

\subsection{Reward design}

From the observation trajectory~$o_0, o_1, \dots$ we score, at each step, the map from~$o_t$ to the stacked window~$(o_{t+1}, \dots, o_{t+\tau})$ of the next~$\tau$ observations, flattened into a single~$d\tau$-dimensional target, using the estimator of Section~\ref{sec:method} with a frozen random four-layer MLP reservoir (ELU nonlinearities). The whole-trajectory score~$S^\phi_T$ is maintained online by the covariance-form recursive-least-squares estimator (Appendix~\ref{app:temporal} gives the recursion), one rank-1 update per step; the per-step reward is the increment~$S^\phi_t - S^\phi_{t-1}$, which telescopes to~$S^\phi_T$.

\subsection{Reinforcement-learning setup}

\emph{Input normalization} is essential here and specific to the reinforcement-learning setting (the inverse experiments use bounded states and do not need it): an agent's observation coordinates are unbounded and can differ in scale across observations by orders of magnitude, so we standardize each coordinate (subtracting its random-policy mean and dividing by its standard deviation) before the reservoir, since a large-magnitude input otherwise saturates the fixed nonlinearity in~$\phi$ and the reward stops discriminating. A \emph{small reservoir} (width~$32$) is the bounded-learner setting: a wide reservoir can fit even a random or chaotic trajectory, so the epiplexity bonus would reward chaos; a narrow one can only fit structured trajectories, so it rewards coherent, regular motion. The \emph{window} horizon~$\tau$ is a fixed per-task constant, on the order of twice the characteristic time over which the state changes appreciably under a random policy, kept within~$[8, 48]$. Per task this gives (state dimension~$d$, horizon~$\tau$): Acrobot ($6$, $8$), MountainCarContinuous ($2$, $28$), Hopper ($11$, $10$), BipedalWalker ($24$, $40$), HalfCheetah ($17$, $16$), LunarLander ($8$, $48$), Walker2d ($17$, $16$), Swimmer ($8$, $16$), Pendulum ($3$, $16$), PointMaze ($8$, $48$). The reservoir ridge is~$\lambda = 0.3$. The PointMaze observation is the flattened goal-conditioned dictionary, and the goal resamples on contact (\texttt{continuing\_task}) so reaching it does not end the episode.

\subsection{Training}

We use PPO~\citep{schulman2017ppo, raffin2021stable} with~$8$ parallel environments, rollout length~$1024$, minibatch~$256$, $\gamma = 0.999$, GAE~$\lambda_{\mathrm{GAE}} = 0.98$, learning rate~$3\times10^{-4}$, for~$600{,}000$ steps, on the default MLP policy, reporting the mean over ten seeds. In the mixed mode the bonus weight~$\beta$ is calibrated per environment so the bonus's whole-episode contribution is~$0.1$ times the random-policy task-return scale (anchored at the episode level so it survives sparse rewards, where a per-step calibration to a near-zero reward would vanish). The state-magnitude control uses the identical setup, with the per-step bonus the squared norm~$\|(o_t - \mu_o) \odot \sigma_o^{-1}\|^2$ of the input-normalized state in place of the epiplexity increment; it is run on all ten tasks of Table~\ref{tab:rl}. The reported return is the task return over~$100$ evaluation episodes, which the epiplexity-only agent never observes during training.

On the sparse and deceptive classic-control tasks the gain is not only in the mean but in reliability: PPO on the task reward alone solves Acrobot and MountainCarContinuous only on some seeds and fails catastrophically on others (across-seed standard deviation~$\approx 166$ and~$43$ in task-return units), whereas the bonus solves them on every seed (standard deviation~$\approx 2$). The bonus turns unreliable exploration into reliable solving. On the locomotion tasks it lifts the mean return (Hopper most, then HalfCheetah, BipedalWalker, and Swimmer), helping PPO escape mediocre local optima. It is rarely harmful: at the same~$600{,}000$-step budget the bonus also raises LunarLander's return (by~$23\%$) and Pendulum's slightly, and only on Walker2d, where the task reward already affords adequate exploration, does it mildly lower the mean (by~$4\%$). The pattern matches the mechanism: the bonus helps exactly when covering observed state is the bottleneck, and the learnable-novelty drive is a mild distraction when it is not.

The state-magnitude control shows that the stability of the bonus is specific to learnable novelty and not shared by raw magnitude. It collapses LunarLander from a return of~$169$ to~$-171$: a near-constant coordinate (the leg-contact indicators, which a random policy almost never triggers, so their estimated standard deviation is tiny) gives the input-normalized magnitude an enormous spike whenever a leg touches down, and the agent is driven toward that spike instead of toward a soft landing. On Walker2d both the magnitude control and the epiplexity bonus sit essentially at the baseline ($294$ and~$285$ against~$296$), neither helping nor collapsing, and on Pendulum the magnitude control leaves the return essentially unchanged. Across all ten tasks the epiplexity bonus falls at most~$4\%$ below the task reward and collapses on none, whereas the magnitude control collapses on two (Hopper and LunarLander). Raw magnitude has no principled scale, so a single degenerate coordinate can hijack it; epiplexity instead prices each coordinate by how well a bounded readout predicts it, which is what makes it stable where the magnitude bonus is not.

\end{document}